\documentclass{article}
\usepackage{PRIMEarxiv}

\usepackage[utf8]{inputenc} 
\usepackage[T1]{fontenc}    
\usepackage{hyperref}       
\usepackage{url}            
\usepackage{booktabs}       
\usepackage{amsfonts}       
\usepackage{nicefrac}       
\usepackage{microtype}      
\usepackage{lipsum}
\usepackage{natbib}
\usepackage{amsmath}
\usepackage{fancyhdr}       
\usepackage{graphicx}       
\graphicspath{{media/}}     
\usepackage{multirow}
\usepackage{algorithm}
\usepackage{algpseudocode}
\usepackage{adjustbox}
\usepackage{tcolorbox}
\usepackage{listings}
\usepackage{xcolor}
\definecolor{promptbg}{RGB}{245,247,250}
\definecolor{promptborder}{RGB}{100,120,180}
\definecolor{fieldbg}{RGB}{235,240,255}
\tcbuselibrary{listings,skins,breakable}

\title{When Planning Problems Become Hard for LLMs: Separating Structural Complexity from Semantic Interpretability
}

\author{
  Mayukh Debnath \\
  Robotics and Autonomous Systems (AI)\\
  Arizona State University\\
  Tempe, USA\\
  \texttt{\{mdebnath\}@asu.edu} \\
   \And
  Adeyinka Adedigba, Rammohan Mallipeddi \\
   Department of Artificial Intelligence\\
  Kyungpook National University \\
  Daegu, South Korea\\
  \texttt{\{yinkpeace, mallipeddi.ram\}@gmail.com} \\
}

\begin{document}
\maketitle

\begin{abstract}
Large language models (LLMs) perform poorly on classical planning, but it is unclear whether this reflects the intrinsic structural complexity of a planning instance or the difficulty of interpreting its symbolic representation. Existing evaluations conflate the two. We study seven STRIPS domains modeled on IPC domains, labeled with breadth-first-search (BFS) optimal plan lengths, and ask two questions. First: can structural difficulty be estimated without solving an instance? A classical, zero-training delete-relaxation heuristic ($h_\text{add}$) estimates optimal plan length on held-out domains far better than every alternative we test, including ARC, a purpose-built episodic retrieval architecture: mean Spearman $|\rho|=0.931$ for $h_\text{add}$, versus $0.707$ for ridge regression on structural features, $0.657$ for object count, and $0.40$ for ARC despite five-seed training with early stopping. ARC does not beat a linear model on its own input features. We take this as the paper's primary finding: for estimating structural planning difficulty, a decades-old AI planning heuristic outperforms a dedicated learned retrieval model, and future work in this space should be judged against it. Second: does an LLM's actual planning success track structural difficulty? We introduce a controlled intervention that progressively renames predicates while preserving the transition system and optimal plans, so structural difficulty is unchanged by construction, and measure planning success with a state-simulating validator: valid-plan success falls from $67.5\%$ to $40.0\%$ as the fraction of renamed symbols grows from $0\%$ to $100\%$, with refusal rate flat at $0\%$ throughout. We further show that an LLM given ARC's own conditioning information estimates difficulty in-context (mean $|\rho|=0.826$) far better than ARC does after training, and that hybrid routing between an LLM and a symbolic planner depends more on calibration with target-model outcomes than on which structural score is used. Together, these results indicate that structural planning difficulty is already well estimated by existing, non-learned methods, and that the open problem for hybrid planning systems is modeling an LLM's semantic, vocabulary-dependent behavior rather than building better structural-difficulty predictors. The complete code is available at \url{https://github.com/mdebnath22/plan}.
\end{abstract}


\section{Introduction}
\label{sec:intro}

Planning - synthesizing goal-directed action sequences over a state space - is a canonical test of systematic reasoning. Large language models (LLMs) remain unreliable at it. \cite{valmeekam-planbench} report that frontier models solve fewer than 3\% of standard PDDL benchmarks; \cite{valmeekam2023planning} show the failures persist under chain-of-thought prompting \cite{wei2022chain} and self-critique \cite{stechly2023gpt4}. Unlike symbolic planners, LLMs perform no explicit combinatorial search, which makes them brittle on problems that require systematically exploring a large state space \cite{kambhampati2024llmmodulo,silver2024generalized}.

Why a given instance is hard is less understood. Two distinct factors can make a planning instance difficult: \emph{structural complexity}, the combinatorial difficulty intrinsic to the state space and goal structure, and \emph{semantic interpretability}, how easily an LLM can ground the symbolic vocabulary describing that same problem. Existing benchmarks conflate them, so a failure cannot be attributed to either cause specifically.

This matters for hybrid planning systems, which route easy instances to an LLM and hard instances to a symbolic planner \citep{kambhampati2024llmmodulo}, use an LLM for heuristic guidance \citep{tang2025think}, or for candidate generation with symbolic verification \citep{valmeekam2023planning}. Every such system needs an upstream difficulty estimate. Existing approaches use problem size, domain identity, or domain-specific classifiers \citep{chiari2025llm}, none of which transfers to a domain with unfamiliar predicates or object vocabulary.

We address this in two parts. First, we ask whether structural difficulty can be estimated without solving an instance, and whether a learned, cross-domain model is the right tool. We build ARC (Asymmetric Retrieval for Cross-domain difficulty estimation), which uses a foundation-model embedding to retrieve structurally similar instances across domains, with structural features providing within-domain discrimination; within a single domain, such embeddings are expected to concentrate in a narrow region of the embedding space, limiting their ability to discriminate difficulty (Section~\ref{sec:collapse}). We compare ARC against object count, goal-fact count, ridge regression on structural features, LogME \citep{you2021logme}, H-score \citep{bao2019hscore}, a non-episodic MLP, and $h_\text{add}$, a delete-relaxation heuristic computed from an instance's grounded action set via a relaxed planning graph, with no learning and no target-domain data. The result is decisive: $h_\text{add}$ reaches mean $|\rho|=0.931$ across three held-out domains; ARC reaches $0.400$, below every baseline except LogME and H-score, and below a linear model fit on exactly its own structural-feature input. We verify $h_\text{add}$ directly: an independent reconstruction of each domain's grounded action set reproduces every saved optimal plan with zero failures across $1{,}006$ solved instances in seven domains.

Second, we ask whether an LLM's actual planning success tracks this structural difficulty. We introduce an intervention that renames a domain's predicates while leaving its transition system, every optimal plan, and every structural feature unchanged. Any change in LLM success under this intervention is attributable to semantic interpretability, not structure. Valid-plan success on Blocksworld falls from $67.5\%$ with no renaming to $40.0\%$ with every predicate renamed, while refusal rate stays at $0\%$ throughout: the model always attempts a plan, and the effect is carried entirely by a rising rate of malformed and inapplicable actions.

\paragraph{Contributions.}
\begin{itemize}
  \item We show that structural difficulty is well estimated by a classical, zero-training delete-relaxation heuristic ($h_\text{add}$, mean $|\rho|=0.931$), which exceeds every alternative we test, including ARC, a retrieval architecture we purpose-built for this problem and evaluated under conditions fixed before training (five seeds, early stopping, a pre-declared criterion for added value). ARC reaches only $0.400$--$0.403$ (ensembled: $0.595$), below a linear model on its own structural-feature inputs ($0.707$). We report this as a substantive negative result for learned cross-domain retrieval, not a limitation of one implementation: Appendix~\ref{app:arc_design} shows the shortfall holds across attention directions, representation combinations, and residualization hyperparameters.
  \item We introduce a semantic intervention based on progressive predicate renaming that holds structural difficulty fixed while varying symbol interpretability. LLM valid-plan success falls from $67.5\%$ to $40.0\%$ as renaming goes from none to complete, with refusal rate flat at $0\%$, demonstrating that semantic interpretability affects LLM planning success independently of structural difficulty.
  \item We show that structure-only routing cannot substitute for target-model calibration, and that an LLM given ARC's own conditioning information outperforms ARC substantially at difficulty estimation (mean $|\rho|=0.826$ vs.\ $0.400$).
\end{itemize}

\section{Related Work}
\label{sec:related}

\paragraph{LLMs for PDDL planning.} LLMs struggle with classical planning, including simple Blocksworld instances \citep{valmeekam2023planning,valmeekam2022large,mcdermott1998pddl}, across IPC-style domains \citep{valmeekam-planbench,ipc2000,ferber2019ipc,silver2020pddlgym}. Self-critique does not reliably fix this \citep{stechly2023gpt4,wei2022chain,guan2023leveraging}. Instruction-tuned PDDL generation improves performance \citep{verma2025pddlinstruct}, but models still generalize poorly to instances larger or structurally different from training \citep{silver2024generalized,tang2025think,tantakoun2025llms,wei2025plangenllms}. We take these limitations as given and ask instead when an LLM should be used at all.

\paragraph{Hybrid planning systems.} The LLM-Modulo framework \citep{kambhampati2024llmmodulo} wraps LLM calls with formal verifiers \citep{gundawar2024robust}. Related work uses LLMs for heuristic guidance \citep{chiari2025llm,correa2026classical} or candidate generation with symbolic verification \citep{li2024systematic,yu2025inspire}. Every such architecture needs an upstream router; existing routing signals do not transfer across domains with unfamiliar vocabulary.

\paragraph{Feature transferability metrics.} LogME \citep{you2021logme} and H-score \citep{bao2019hscore} estimate how well a frozen feature space supports a downstream task without fine-tuning \citep{bao2018information,nguyen2020leep,sae2025transferability}, typically by fitting a linear-Gaussian model and reporting its marginal likelihood or an information-theoretic divergence. We use both as zero-shot baselines, fit on source-domain instances and scored on held-out domains.

\paragraph{Meta-learning.} MAML \citep{finn2017maml} performs an explicit inner-loop update per episode; Prototypical Networks \citep{snell2017prototypical} score a query by its relation to a labeled support set with no inner-loop adaptation. ARC follows the latter: one set of parameters, updated once per episode, with no meta-/task-parameter distinction. ``Episodic'' here refers only to the sampling procedure \citep{blanchard2011generalizing,blanchard2021domain}: each step draws one source domain and conditions on a labeled support set from it, drawing on the geometric structure of PDDL embeddings \citep{reimers2019sbert,oord2018cpc}.

\section{Background and Problem Formulation}
\label{sec:background}

\paragraph{PDDL and planning instances.} A classical planning instance is a pair $\pi = \langle D, P\rangle$, where $D$ specifies a domain's predicates and lifted operators in PDDL, and $P = \langle O, s_0, g\rangle$ gives a set of objects $O$, an initial state $s_0$, and a goal condition $g$. A plan is a sequence of ground actions $\mathbf{a} = (a_1,\dots,a_L)$ whose execution from $s_0$ satisfies $g$. The optimal plan length is
\begin{equation}
    n^\star(\pi) = \min_{\mathbf{a}} \left\{ |\mathbf{a}| : g \subseteq \Gamma(s_0, \mathbf{a}) \right\},
\end{equation}
where $\Gamma(s_0,\mathbf{a})$ is the state reached by executing $\mathbf{a}$ from $s_0$. We take $n^\star(\pi)$ as our operational definition of structural difficulty.

\begin{tcolorbox}[
  enhanced, breakable,
  colback={rgb,255:red,247;green,249;blue,252},
  colframe={rgb,255:red,120;green,130;blue,160},
  fonttitle=\bfseries\small,
  title={Example: A Blocksworld planning instance},
  top=6pt, bottom=6pt, left=6pt, right=6pt
]
\small
Three blocks $a$, $b$, $c$, initially on the table and clear. Goal: stack $a$ on $b$.
\begin{lstlisting}[basicstyle=\ttfamily\footnotesize]
(:objects a b c)
(:init (on-table a) (on-table b) (on-table c)
       (clear a) (clear b) (clear c) (handempty))
(:goal (on a b))
\end{lstlisting}
Here $O=\{a,b,c\}$, $s_0$ is the \texttt{init} block, $g=\{(\text{on } a\ b)\}$. One action, $\text{stack}(a,b)$, reaches a state satisfying $g$, so $n^\star(\pi)=1$. Instances in our evaluation are larger; this minimal example fixes notation.
\end{tcolorbox}

\paragraph{Difficulty labels.} We label every instance by breadth-first search (BFS) within a budget $B=(T_\text{max}{=}12, N_\text{max}{=}5{,}000)$, identically across domains: the regression target is $y_r=n^\star(\pi)$. Instances BFS does not solve within $B$ carry no label and are excluded from every computation in this paper; a finite-budget failure neither establishes unsolvability nor yields a plan-length value, so we never assign one. Per-domain coverage is in Appendix~\ref{app:label_availability}.

\paragraph{Feature representation.} Each instance has three views. \textbf{Structural features} $x_s\in\mathbb{R}^{28}$ --- object count, initial- and goal-fact counts and ratios, predicate-arity statistics, operator count --- contain no domain identifier and nothing derived from the plan or $n^\star$. \textbf{Foundation-model embeddings} $x_e\in\mathbb{R}^{768}$ encode a natural-language description built from object counts and a plain-language summary of the initial state and goal, with no raw PDDL syntax and nothing derived from the plan; no description contains a step count or difficulty word. \textbf{The residual} $x_r=x_e-\hat{x}_e(x_s)$ is the component of $x_e$ not explained by $x_s$, estimated by PCA-Ridge regression. All preprocessing is fit on pooled source-domain data and frozen before evaluation.

\paragraph{FM embedding collapse.}
\label{sec:collapse}
ARC rests on a diagnostic hypothesis: $x_e$ should identify semantic similarity across domains but concentrate within a domain, limiting within-domain discrimination; $x_s$ should do the opposite. We embed every instance's description with \texttt{all-mpnet-base-v2} and measure mean within-domain pairwise cosine similarity and effective rank $\exp(\mathcal{H}(\sigma/\|\sigma\|_1))$, $\sigma$ the embedding matrix's singular values. On the three held-out domains, mean cosine similarity is $0.989$/$0.986$/$0.992$ (BW/LOG/MBW), effective rank $2.62$; two further encoders give $0.988$/$0.988$ cosine and $2.64$/$2.61$ rank (Appendix~\ref{app:encoder_robustness}), so this is a property of the encoder family on these descriptions, not one model.

Structural features, pooled-standardized, range $0.69$--$0.95$ cosine and $2.1$--$5.3$ effective rank (mean $3.8$) within a domain. The hypothesized asymmetry holds in direction --- $x_e$ is consistently more collapsed than $x_s$ --- but the effective-rank gap is smaller than the cosine gap suggests: $x_s$ is also concentrated within a domain, not uniformly higher-dimensional. Whether the asymmetric query/key assignment this motivates is in fact the best way to exploit this structure is answered empirically in Appendix~\ref{app:arc_design}: on balance, no --- it is the best of four tested configurations (Appendix~\ref{app:attention_direction}), but none is competitive with the non-episodic baselines in Table~\ref{tab:main_results}.

\begin{figure*}[t]
  \centering
  \includegraphics[width=\linewidth]{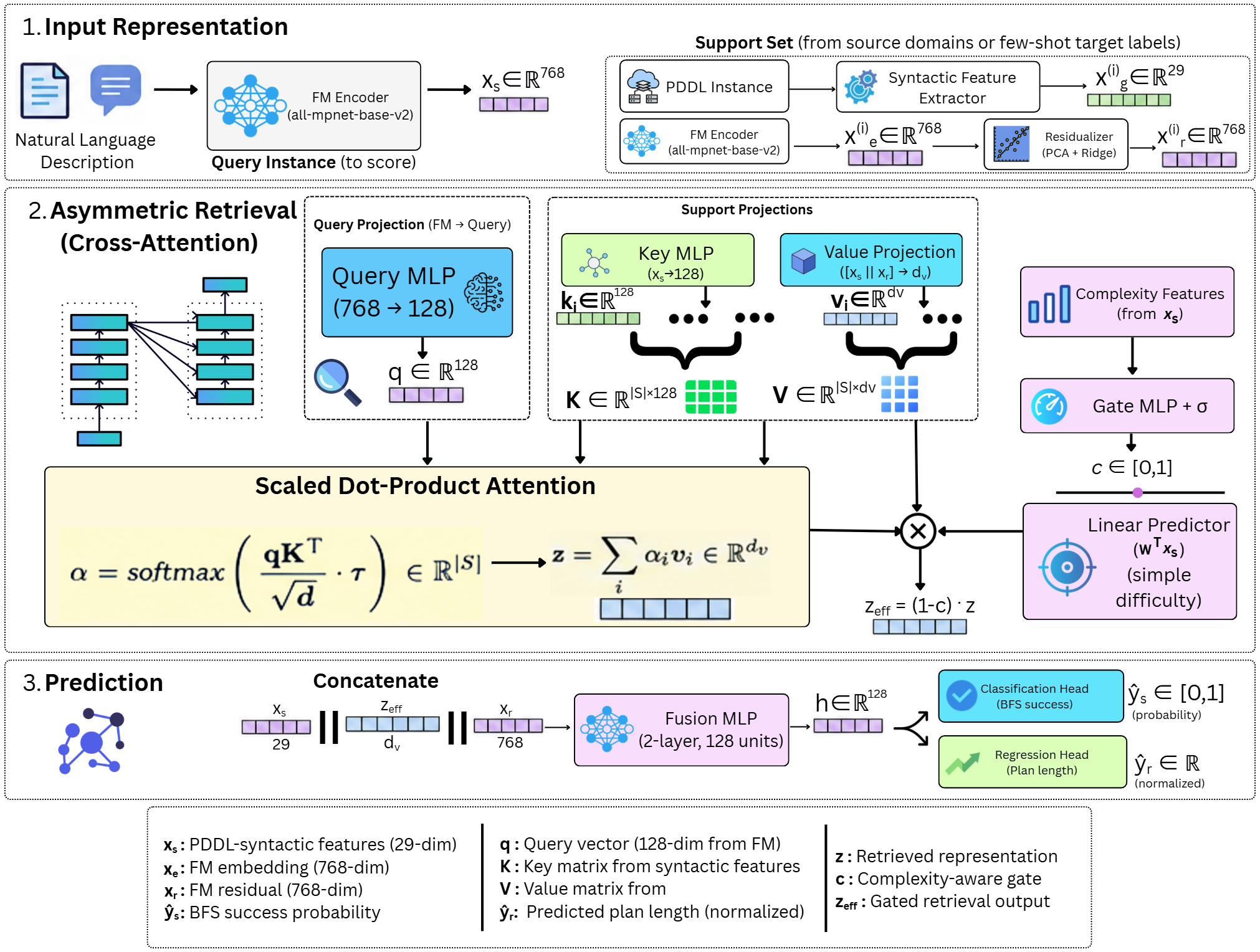}
  \caption{ARC architecture. A query instance's FM embedding is projected to a 128-dimensional query vector via QueryMLP. Support instances' structural features form keys (KeyMLP); a concatenation of structural and residual FM features forms values. Soft attention retrieves a weighted difficulty summary $\mathbf{z}$. The fusion head concatenates $\mathbf{x}_s^q$, $\mathbf{z}$ and $\mathbf{x}_r^q$ and produces a regression output (normalized plan length). An optional gate blending the attention output with a linear predictor is evaluated as a variant (Section~\ref{sec:arc}).}
  \label{fig:arc_arch}
\end{figure*}

\section{ARC: Architecture and Training}
\label{sec:arc}

ARC estimates difficulty by retrieving structural analogues of a query instance from a labeled support set (60 instances, at training and test time). No target-domain instance, label, or statistic is used at test time. The semantic embedding of the query retrieves structurally relevant support instances: $x_e$ aligns across domains as a query; $x_s$ discriminates within a domain as keys and values. Algorithm~\ref{alg:arc_inference} gives inference.

\begin{algorithm}[h]
\caption{ARC Inference (Asymmetric Retrieval)}
\label{alg:arc_inference}
\begin{algorithmic}[1]
\Require Query $x_e^q, x_s^q, x_r^q$; support set $\{(x_s^{(i)}, x_r^{(i)}, y^{(i)})\}_{i=1}^{|\mathcal{S}|}$
\Ensure Predicted normalized plan length $\hat{y}_r$
\State $q \gets W_{Q_2}\cdot\text{GELU}(\text{LN}(W_{Q_1} x_e^q))$
\For{$i=1,\dots,|\mathcal{S}|$}
  \State $k_i \gets W_{K_2}\cdot\text{GELU}(W_{K_1} x_s^{(i)})$; \quad $V_i \gets \text{ValueProj}([x_s^{(i)}\Vert x_r^{(i)}])$
\EndFor
\State $\alpha \gets \text{softmax}(\tau\cdot qK^\top/\sqrt{d})$; \quad $z \gets \alpha^\top V$
\State $h \gets \text{FusionMLP}([x_s^q \Vert z \Vert x_r^q])$; \quad $\hat{y}_\text{attn} \gets \text{RegressionHead}(h)$
\If{gated variant} $c \gets \sigma(\text{MLP}(x_s^q))$; \quad $\hat{y}_r \gets c\cdot w^\top x_s^q + (1-c)\hat{y}_\text{attn}$
\Else\ \ $\hat{y}_r \gets \hat{y}_\text{attn}$
\EndIf
\State \Return $\hat{y}_r$
\end{algorithmic}
\end{algorithm}

The query encoder maps $x_e$ to a 128-dimensional space through a two-layer MLP (LayerNorm, dropout $0.1$); the key encoder does the same for $x_s$. The value projection concatenates $x_s^{(i)}$ and $x_r^{(i)}$, so retrieval carries both structural and residual semantic information. Attention temperature starts at $\tau=1$.

\paragraph{Optional complexity-aware gate.} We also evaluate, as a pre-declared variant, a gate $c=\sigma(\text{MLP}(x_s^q))\in[0,1]$ blending a linear predictor $\hat{y}_\text{simple}=w^\top x_s^q$ with the attention output, trained end-to-end with no auxiliary supervision. The main configuration below uses no gate; the gated variant is reported alongside it, and we do not select between them using test-domain scores.

\paragraph{Budget-exceeded instances.}
\label{sec:budget_exceeded}
An instance BFS does not solve within its budget is not known to be unsolvable, nor necessarily harder than a solved one. Such instances carry no label and supervise no head of the model.

\subsection{Training Objective}
\label{sec:training_objective}
ARC trains for up to 5{,}000 episodes. Each episode samples one source domain, splits its labeled instances into a 60-instance support set and 30-instance query set, and minimizes
\begin{equation}
  \mathcal{L} = \mathcal{L}_\text{reg} - \lambda_e\,\mathcal{H}(\boldsymbol{\alpha}),
\end{equation}
mean squared error between the prediction and the domain-level z-scored plan length, regularized by the negative entropy of the attention weights ($\lambda_e=0.05$) to prevent collapse onto one support instance. Every term is optimized by one outer AdamW update (lr $3\times10^{-4}$, weight decay $10^{-4}$, cosine annealing to $10^{-5}$, gradient clipping $1.0$) per episode - there is no inner-loop adaptation step. Gripper is the validation domain: every 200 episodes we score ARC zero-shot on Gripper, keep the best checkpoint, and stop after 10 checks with no improvement; the kept, not the last, checkpoint is evaluated. We report mean $\pm$ std over five seeds (0, 1, 2, 42, 123) for every trained method.

\section{Experiments}
\label{sec:experiments}

\subsection{Setup}
\label{sec:setup}

\paragraph{Domains.} Seven classical STRIPS domains modeled on IPC domains, generated by our own code: source domains Depot, Rovers, Satellite; held-out test domains Blocksworld (BW), Logistics (LOG), Mystery-Blocksworld (MBW); validation domain Gripper (early stopping only). Each has 200 generated instances; only BFS-solved instances carry a label and enter any computation: 126 (BW), 164 (LOG), 133 (MBW), 132 (Depot), 199 (Rovers), 126 (Satellite), 181 (Gripper). Mystery-Blocksworld shares Blocksworld's state space but replaces every predicate and action name with an uninformative symbol.

\paragraph{Metric.} Our primary metric is absolute Spearman rank correlation $|\rho|$ between a predicted score and $n^\star$, over each test domain's BFS-solved instances; a ranking measure requiring no threshold tuning. For routing (Section~\ref{sec:routing}), we report the fraction of instances solved, with plan validity checked by a state-simulating executor. Confidence intervals: Appendix~\ref{app:primary_ci}.

\paragraph{Baselines.} Object count $|\mathcal{O}|$; unsatisfied goal-fact count; ridge regression on $x_s$; LogME, fit with evidence-optimal regularization rather than a fixed value; H-score; a non-episodic MLP (ERM); and $h_\text{add}$, computed from each instance's grounded action set via a relaxed planning graph with all delete effects ignored, with no learning and no target-domain data. $h_\text{add}$ is not an $O(1)$ read: it needs the same grounded action set a planner would use. We reconstruct this independently per domain --- from object-naming conventions and each instance's own saved initial state, never reusing the labeling code --- and verify it: re-simulating every saved BFS-optimal plan against the reconstructed action set succeeds with zero failures across all $1{,}006$ labeled instances in seven domains.

\section{Results}

\subsection{Zero-Shot Estimation of Structural Complexity}
\label{sec:structural_estimation}

\begin{table}[t]
\centering
\begin{tabular}{lcccc}
\toprule
\textbf{Method} & \textbf{BW} & \textbf{LOG} & \textbf{MBW} & \textbf{Mean} \\
\midrule
Object count ($|\mathcal{O}|$)        & 0.764 & 0.544 & 0.662 & 0.657 \\
Goal count                            & 0.764 & 0.969 & 0.673 & 0.802 \\
Ridge on $x_s$                        & 0.644 & 0.889 & 0.587 & 0.707 \\
Ridge on $x_s$ (z-scored target)      & 0.645 & 0.889 & 0.589 & 0.708 \\
LogME                                 & 0.057 & 0.704 & 0.478 & 0.413 \\
H-score                               & 0.648 & 0.180 & 0.336 & 0.388 \\
ERM (non-episodic)                    & 0.537$\pm$0.018 & 0.865$\pm$0.043 & 0.507$\pm$0.018 & 0.636 \\
ERM (z-scored target)                 & 0.537$\pm$0.027 & 0.884$\pm$0.014 & 0.482$\pm$0.015 & 0.634 \\
\midrule
\textbf{Delta-relaxation heuristic ($h_\text{add}$)} & \textbf{0.894} & \textbf{0.976} & \textbf{0.923} & \textbf{0.931} \\
\midrule
ARC (main configuration)              & 0.294$\pm$0.216 & 0.652$\pm$0.297 & 0.254$\pm$0.157 & 0.400 \\
ARC + blend gate                      & 0.292$\pm$0.181 & 0.668$\pm$0.325 & 0.248$\pm$0.146 & 0.403 \\
ARC + blend gate, prediction-ensembled & 0.482 & 0.912 & 0.390 & 0.595 \\
\bottomrule
\end{tabular}%
\caption{Absolute Spearman correlations between predicted difficulty scores and optimal plan lengths on held-out domains ($N=126$, $164$, $133$ BFS-solved instances for BW, LOG, MBW). ERM and ARC rows report mean $\pm$ std over five seeds; the ensembled ARC row averages the five seeds' predictions before computing one correlation. Bold marks the overall best row. LogME uses evidence-optimal regularization ($\alpha/\beta=2.358$) rather than a fixed value.}
\label{tab:main_results}
\end{table}

\textbf{$h_\text{add}$ transfers best, and no scalar statistic dominates every domain.} $h_\text{add}$ reaches mean $|\rho|=0.931$ ($0.894$/$0.976$/$0.923$). Goal count reaches $0.802$, ridge on $x_s$ reaches $0.707$, object count $0.657$. A paired bootstrap ($2{,}000$ resamples) of $h_\text{add}$ against goal count gives $\Delta|\rho|=+0.119$ $[+0.057,+0.192]$ on BW and $+0.236$ $[+0.143,+0.344]$ on MBW; on Logistics the two are indistinguishable ($\Delta=0.000$), because $h_\text{add}$ reduces to a monotonic function of unsatisfied-goal count on this domain's relaxed graph (Appendix~\ref{app:cheap_baselines}). Object count exceeds ridge on BW and MBW; ridge is far stronger on LOG ($0.889$ vs.\ $0.544$). $h_\text{add}$ is not an $O(1)$ read, so we verify its action-set reconstruction directly: zero failures re-simulating $1{,}006$ saved plans. Appending $h_\text{add}$ as a 29th feature to ridge on $x_s$ is mixed, not uniform ($+0.095$ MBW, $-0.001$ LOG, $-0.077$ BW), so its standalone strength is not simply additive under an untuned fit.

\textbf{ARC does not add value beyond these baselines.} We fixed this criterion in advance: added value requires exceeding the strongest baseline in every comparison by more than the seed-to-seed standard deviation. It does not. ARC reaches mean $0.400$ ($0.294\pm0.216$/$0.652\pm0.297$/$0.254\pm0.157$); the gated variant, $0.403$. Both trail every deterministic baseline except LogME and H-score, and $h_\text{add}$ by a wide margin. The standard deviations are large relative to the means ($\pm0.216$ against a mean of $0.294$ on BW), so single-seed ARC numbers are unreliable estimates of transfer performance. Averaging predictions across seeds before correlating, rather than averaging correlations, partially closes the gap (mean $0.595$: $0.482$/$0.912$/$0.390$) but still falls short of $h_\text{add}$ and of ridge on ARC's own input. ARC uses only PDDL-derived features and a labeled support set, with no target-domain labels or updates at test time --- yet a linear model using strictly less of that information outperforms it. Appendix~\ref{app:arc_design} tests whether this is specific to one design choice (attention direction, feature combination, residualization) and finds the shortfall persists throughout.

\begin{table}[h]
\centering
\begin{tabular}{lcccccc}
\toprule
\textbf{Domain} & \textbf{B1} & \textbf{B2} & \textbf{B3} & \textbf{B4} & \textbf{B5} & \textbf{Mean} \\
\midrule
Ridge on $x_s$, BW  & 0.48 & 0.37 & 0.58 & 0.45 & 0.57 & 0.49 \\
Ridge on $x_s$, LOG & 0.87 & 0.96 & 0.96 & 0.98 & n/a  & \textbf{0.94} \\
Ridge on $x_s$, MBW & 0.07 & 0.27 & 0.24 & 0.10 & 0.19 & 0.17 \\
\midrule
ARC (main), BW  & 0.11 & 0.32 & 0.22 & 0.20 & 0.36 & 0.24 \\
ARC (main), LOG & 0.23 & 0.36 & 0.33 & 0.33 & n/a  & 0.32 \\
ARC (main), MBW & 0.15 & 0.19 & 0.16 & 0.17 & 0.27 & 0.19 \\
\midrule
$|\mathcal{O}|$ & 0.00 & 0.00 & 0.00 & 0.00 & 0.00 & 0.00 \\
\bottomrule
\end{tabular}%
\caption{Absolute Spearman correlation within fixed object-count groups. B1--B5 are the five generator size levels (object counts $2,3,4,6,8$ for BW/MBW and $4,5,6,7,9$ for LOG). Group sizes are $N=40,40,24,13,9$ (BW), $40,40,40,40,4$ (LOG; B5 undefined, two distinct plan lengths), $40,40,28,14,11$ (MBW); small groups are noisy. Object count is constant within a group, so its baseline has zero ranking capacity.}
\label{tab:within_size}
\end{table}

\textbf{A coarse-proxy explanation does not hold.} Within fixed object-count groups, where $|\mathcal{O}|$ carries no ranking information, ridge on $x_s$ keeps mean $|\rho|=0.49$/$0.94$/$0.17$ (BW/LOG/MBW; Table~\ref{tab:within_size}) --- real within-group signal, strongly on LOG, weakly on MBW; we do not have an explanation for the gap between the two Blocksworld variants, which share a state space. ARC's within-group correlations are weaker still ($0.24$/$0.32$/$0.19$), trailing ridge in all three domains. Which cheap statistic wins also depends on the domain: goal count and ridge dominate object count on LOG, while object count dominates ridge on BW and MBW --- full breakdown in Appendix~\ref{app:within_size}.

\textbf{Robustness.} The baseline ordering is not an artifact of one split: in leave-one-domain-out over all seven domains, ridge exceeds object count on six of seven (mean $0.732$ vs.\ $0.520$), and ARC, retrained per fold, trails ridge on all seven (mean $0.640$ vs.\ $0.732$; Appendix~\ref{app:loo_generalisation}). Excluding the few BFS-solved instances with $n^\star=0$ changes the means to $0.813$/$0.809$/$0.643$/$0.931$ (goal count/ridge/object count/$h_\text{add}$, the last unchanged to three decimals). Solver-effort association, the LLM-based difficulty baseline, and three custom non-IPC domains extend this picture further (Appendices~\ref{app:solver_association}, \ref{app:llm_baseline}, \ref{app:custom_datasets}).

\subsection{Semantic Perturbations Separate Structural from Behavioral Difficulty}
\label{sec:structural_behavioral}

Section~\ref{sec:structural_estimation} asked whether difficulty can be \emph{predicted} from structure; here we ask whether an LLM's \emph{behavioral} success tracks that same difficulty, or moves independently of it when only the problem's symbols change. Qwen~2.5-72B is the system under study here --- generating and attempting plans, not predicting difficulty (that role is evaluated separately in Appendix~\ref{app:llm_baseline}).

We progressively corrupt the same 40 BFS-solved Blocksworld instances. For each $k\in\{0,25,50,75,100\}\%$, we replace a fraction $k$ of predicate and action names with meaningless tokens, applied consistently across domain and problem files. This preserves every precondition, effect, state, goal, and optimal plan length $n^\star(\pi)$; only the vocabulary's readability changes. A plan is valid only if every action uses a name from the renamed domain shown to the model and, mapped back, passes a state-simulating executor (parsing, arity, preconditions, goal; Appendix~\ref{app:llm_protocol}).

\begin{table}[h]
\centering
\small
\begin{tabular}{cccc}
\toprule
\textbf{Predicate Corruption (\(k\))} & \textbf{Valid-Plan Success} & \textbf{95\% Wilson CI} & \textbf{Refusal Rate} \\
\midrule
0\%   & 67.5\% & [52.0\%, 79.9\%] & 0.0\% \\
25\%  & 62.5\% & [47.0\%, 75.8\%] & 0.0\% \\
50\%  & 47.5\% & [32.9\%, 62.5\%] & 0.0\% \\
75\%  & 37.5\% & [24.2\%, 53.0\%] & 0.0\% \\
100\% & 40.0\% & [26.3\%, 55.4\%] & 0.0\% \\
\bottomrule
\end{tabular}
\caption{Qwen~2.5-72B performance on progressively corrupted versions of the same 40 BFS-solved Blocksworld instances, validated by a state-simulating executor. Renaming preserves the transition system and optimal plan lengths \(n^\star(\pi)\).}
\label{tab:corruption}
\end{table}

Success falls from $67.5\%$ to $40.0\%$ (Table~\ref{tab:corruption}); the endpoint intervals, $[52.0,79.9]$ and $[26.3,55.4]$, barely overlap. The decline is not perfectly monotonic step to step ($47.5\to37.5\to40.0$ across the last three levels), consistent with $N=40$ being too small to resolve small adjacent differences; we read the endpoint-to-endpoint effect, not a strict trend. Refusal rate is $0\%$ at every level --- the model always attempts a plan, and the effect is carried entirely by a rising rate of malformed and inapplicable actions.

Structural difficulty is unchanged across $k$ by construction. Comparing structural estimators across BW and MBW in Table~\ref{tab:main_results} is \emph{not} a paired test of this, since they are independently generated instance sets, not renamings of the same instances (goal count $0.764$ vs.\ $0.673$; ridge $0.644$ vs.\ $0.587$; object count $0.764$ vs.\ $0.662$). The paired test is the scrambling experiment itself, where $n^\star(\pi)$ is identical for every instance at every $k$: the $27.5$-point drop, with no accompanying refusal-rate change, is attributable entirely to semantic interpretability. The same transition system, the same optimal plan length, the same object count produce a substantially different success rate purely as a function of vocabulary interpretability.

\subsection{Structural Difficulty Alone Is Insufficient for LLM Routing}
\label{sec:routing}

Each instance is routed to Qwen~2.5-72B or a symbolic fallback; a routed instance counts as solved only if the selected component returns a valid plan. We route structurally easy instances to the LLM, motivated by LLM failure rates rising sharply with plan length \citep{valmeekam2023planning}, while the symbolic solver is guaranteed correct whenever a plan exists, at the cost of speed. Static always uses the fallback; Oracle picks the successful component per instance. Zero-shot (ZS) policies threshold a structural score alone; few-shot (FS) policies additionally calibrate on 140 target-domain LLM outcomes. We include goal count as the strongest $O(1)$ statistic in Table~\ref{tab:main_results}; $h_\text{add}$ would likely be stronger still, which we flag as a natural extension rather than one we have run.

\begin{table*}[t]
\centering
\small
\begin{tabular}{lcccccccc}
\toprule
\textbf{Domain} & \textbf{Static} & \textbf{Oracle} & $|\mathcal{O}|$-ZS & Goal-ZS & \textbf{ARC-ZS} & $|\mathcal{O}|$-FS & Goal-FS & \textbf{ARC-FS} \\
\midrule
Blocksworld (BW)       & 100.0 & 100.0 & 64.0 & 64.0 & 73.0  & 86.0 & 64.0 & 99.0  \\
Logistics (LOG)        & 100.0 & 100.0 & 32.0 & 23.0 & 50.0  & 77.0 & 59.0 & 99.0  \\
Mystery-BW (MBW)       & 100.0 & 100.0 & 64.0 & 64.0 & 65.0  & 85.0 & 64.0 & 100.0 \\
\bottomrule
\end{tabular}
\caption{Percentage of 100 sampled instances per domain solved under each routing policy, validated by the state-simulating executor. ZS uses a structural score with no target-domain LLM outcomes; FS calibrates the decision threshold on 140 target-domain outcomes. Static and Oracle are both $100\%$ because the symbolic fallback always succeeds on these BFS-solvable instances (no time or resource budget is imposed).}
\label{tab:routing}
\end{table*}

\textbf{What Static$=$Oracle$=100\%$ means.} The fallback always succeeds, so no policy can exceed $100\%$, and routing to the LLM can only match or risk that outcome, not improve it; this table does not measure speed or cost. We read it instead as asking: among instances routed to the LLM, how often would it also have succeeded? A comparison between scoring methods, not between routing and never-routing.

\textbf{ARC-FS is the strongest routing signal despite being the weakest ranker.} ARC-FS reaches $99$--$100\%$ on every domain, ahead of $|\mathcal{O}|$-FS ($86.0$/$77.0$/$85.0$) and Goal-FS ($64.0$ everywhere); ARC-ZS is the best zero-shot signal on BW and LOG, tied on MBW. This is notable because ARC was the \emph{weakest} ranker in Section~\ref{sec:structural_estimation}. Ranking requires correct order across the full difficulty range; routing requires only separating ``likely succeeds'' from ``likely fails'' at one threshold, a coarser distinction a poor ranker can apparently still support. We do not read this as rehabilitating ARC as an estimator --- routing performance should not be assumed to track ranking performance, in either direction.

LOG is where every zero-shot policy is weakest ($\leq50\%$) and where calibration helps most (ARC: $+49$ points ZS$\to$FS), consistent with the LLM's own LOG success being near zero regardless of structural difficulty (Appendix~\ref{app:llm_capability_extended}): little for a zero-shot score to find, so calibration does most of the work of learning where the LLM can be trusted. A router calibrated on a small amount of target-model feedback adapts to vocabulary-specific LLM behavior far more than a zero-shot structural score does - using a score that is, in isolation, a poor ranker. Practical hybrid systems should combine a general structural score with behavioral feedback rather than relying on either alone: zero-shot scores here are modest (at best $73\%$ correctly identified as LLM-tractable), while 140 calibration examples close most of the remaining gap for every method tested.

\section{Conclusion}
\label{sec:conclusion}

We asked whether the structural difficulty of a classical planning instance can be estimated without solving it, and whether an LLM's actual planning success tracks that difficulty. On seven STRIPS domains, structural difficulty is well estimated by simple, training-free means: $h_\text{add}$ reaches mean $|\rho|=0.931$ with optimal plan length, goal count $0.802$, ridge on 28 structural features $0.707$, object count $0.657$. Ridge beats object count on six of seven domains leave-one-domain-out, and $h_\text{add}$'s correctness is independently verified: zero failures re-simulating $1{,}006$ saved BFS plans against freshly reconstructed action sets.

The learned-estimation result is squarely negative. ARC, built specifically to combine structural features with foundation-model embeddings under a protocol fixed before training (five seeds, early stopping, a pre-declared added-value criterion), does not meet that criterion: mean $0.400$ ($0.403$ gated), below every deterministic baseline except LogME and H-score, below a linear model on its own inputs ($0.707$), below its own weaker input scored alone ($x_e$ alone: $0.412$), far below $h_\text{add}$. The shortfall persists leave-one-domain-out, within object-count groups, on three custom domains outside the design loop, and across every attention-direction and residualization variant tested, including one post hoc setting we decline to adopt because selecting it required looking at the test domains. Prediction-level ensembling partially closes the gap ($0.595$) but does not close it. We read this as a genuine negative result for cross-domain learned retrieval on this problem: a classical, zero-training heuristic transferring better than a purpose-built learned architecture is itself a useful finding, and future estimators in this space should be reported alongside $h_\text{add}$, not only against weaker scalar baselines.

A controlled predicate-renaming intervention holds structural difficulty fixed by construction while varying vocabulary interpretability: valid-plan success falls from $67.5\%$ to $40.0\%$ as corruption rises from $0\%$ to $100\%$, refusal rate flat at $0\%$ throughout, the entire effect carried by malformed and inapplicable actions rather than the model declining to answer. This separates \emph{structural difficulty}, already well estimated by $h_\text{add}$, from \emph{behavioral difficulty}, which depends on semantic interpretability and is not reducible to it. The consequence for hybrid planning is more nuanced than ``routing helps'' or ``routing fails'': with no solver-failure budget modeled, every policy is bounded above by always-use-the-solver, so routing instead measures which structural score best identifies LLM-tractable instances --- and a calibrated ARC score is the best such signal despite being the weakest ranker in the paper, indicating routing and ranking are distinct tasks. An LLM given ARC's own conditioning information, with no training, estimates difficulty via in-context reasoning (mean $|\rho|=0.826$) far better than ARC does after training, second only to $h_\text{add}$ overall.

Structural planning difficulty across these domains is close to a solved estimation problem: a 1970s-vintage AI-planning technique solves it better than the learned architecture we purpose-built for the task, and an LLM's untrained in-context reasoning over the same inputs matches that technique closely. The open problem for hybrid LLM-planning systems is not a better structural-difficulty predictor, but modeling the semantic, vocabulary-dependent component of LLM behavior that a structural score, by construction, cannot see, and that routing cannot fully compensate for even when calibrated.

\bibliographystyle{unsrt}  
\bibliography{references}

\newpage

\appendix
\appendix

\section{Data, Labels, and Evaluation Protocol}
\label{app:data_protocol}
This appendix clarifies the supervision used to train ARC and the protocol used to evaluate LLM-generated plans.
\subsection{Label Availability and Supervision}
\label{app:label_availability}
Each planning domain contains \(N=200\) generated instances (40 at each of five size levels). ARC is trained using a single supervision signal derived from BFS: the optimal plan length,
\[
y_r(\pi)=n^\star(\pi),
\]
used to learn a continuous structural-difficulty ranking. This label is available whenever breadth-first search returns a solution under the budget \(B=(T_{\max}=12,N_{\max}=5{,}000)\), applied identically to every domain. An instance for which the search budget is exhausted is excluded from training and from all rank-based evaluation, because failure under a finite budget neither establishes unsolvability nor yields a plan-length value; such instances are never assigned a substitute label, and no head of the model is supervised with them.

\begin{table}[h]
\centering
\small
\begin{tabular}{llrrrcr}
\toprule
\textbf{Domain} & \textbf{Role} & \textbf{BFS-solved} & \textbf{Solved \%} & $|\mathcal{O}|$ & $n^\star$ (min/med/max) & $n^\star{=}0$ \\
\midrule
Blocksworld (BW)       & test  & 126 & 63.0 & 2--8  & 0/4/10 & 15 \\
Logistics (LOG)        & test  & 164 & 82.0 & 4--9  & 0/5/12 & 6 \\
Mystery-BW (MBW)       & test  & 133 & 66.5 & 2--8  & 0/4/10 & 10 \\
Depot                  & train & 132 & 66.0 & 4--7  & 7/8/10 & 0 \\
Rovers                 & train & 199 & 99.5 & 4--15 & 2/3/5  & 0 \\
Satellite              & train & 126 & 63.0 & 4--12 & 3/5/7  & 0 \\
Gripper                & val.  & 181 & 90.5 & 4--12 & 0/5/10 & 24 \\
\bottomrule
\end{tabular}
\caption{Per-domain data statistics (200 generated instances per domain). ``BFS-solved'' is the number of instances for which BFS finds a plan within $B$; only these carry a label and enter any training or evaluation. The last column counts BFS-solved instances whose goal already holds in the initial state.}
\label{tab:dataset_domains}
\end{table}

Coverage of the label is incomplete and domain-dependent (Table~\ref{tab:dataset_domains}): between $63.0\%$ (Blocksworld, Satellite) and $99.5\%$ (Rovers) of generated instances are BFS-solved; the remainder are excluded from every computation. The rank-based correlations reported throughout the main text are evaluated only on instances that possess a valid plan-length target. Two properties of the labels deserve mention. First, the source domains Depot ($n^\star\in[7,10]$) and Rovers ($n^\star\in[2,5]$) cover narrow ranges of plan length. Second, a small number of BFS-solved instances have $n^\star=0$ (the goal already holds initially; 15 in BW, 6 in LOG, 10 in MBW). They are included in all primary results; excluding them changes the test-domain means to $0.813$ (goal count), $0.809$ (ridge on $x_s$), and $0.643$ (object count), with $N=111$, $158$, $123$ remaining instances for BW, LOG, MBW. All preprocessing transformations, including feature standardisation, PCA, and residualisation, are fitted exclusively on pooled source-domain instances and frozen before evaluation on the held-out domains.

Because BFS is capped at a fixed budget, the hardest instances in each domain's difficulty distribution are systematically more likely to be excluded from the label. The correlations reported throughout this paper should therefore be interpreted as characterizing the difficulty ordering within the search-feasible regime, and may not extend uniformly into the harder, budget-exceeding tail of each domain.

\subsection{LLM Prompt and Plan-Validity Protocol}
\label{app:llm_protocol}
We evaluate Qwen~2.5-72B on the 200 generated instances of each held-out domain for the routing experiment, and on 40 BFS-solved Blocksworld instances at each corruption level for the scrambling experiment. Generation uses temperature \(0\), a maximum of \(2{,}048\) output tokens, and no domain-specific demonstrations. Each prompt contains the complete PDDL domain file and the problem file for one instance. The model is instructed to reason through the state transitions and to return only the final action sequence. The prompt template is reproduced in full in Appendix~\ref{app:prompts}.

Generated plans are evaluated by execution rather than by textual similarity to a reference plan. A custom PDDL state-space executor performs three checks: (1)~every output line must parse as a syntactically valid grounded action; (2)~each action's preconditions must hold in the state produced by all preceding actions; and (3)~the goal condition must hold after the final action is applied. A response is counted as a valid plan only if all three checks succeed. Any malformed action, inapplicable action, or final state that fails to satisfy the goal is counted as a planning failure; no partial credit is assigned. For renamed domains, an action name that does not belong to the renamed domain shown to the model is also counted as a failure, and the remaining actions are mapped back to the original names before the executor is applied (renaming preserves the transition system).

For the semantic-corruption analysis, a refusal is recorded when the model returns the designated NOPLAN response or produces an empty response. Refusals are reported separately from invalid generated plans: the former indicate that no executable action sequence was proposed, whereas the latter indicate that the model attempted a plan that failed formal verification.

\subsection{LLM Capability Threshold Across Model Families}
\label{app:llm_capability_extended}

To assess whether the qualitative findings of Section~\ref{sec:structural_behavioral} are specific to Qwen~2.5-72B or reflect a more general property of current open-weight LLMs, we evaluate the same BW/LOG/MBW protocol (Appendix~\ref{app:llm_protocol}) on three additional models spanning a $10\times$ parameter range and two model families distinct from Qwen: Mistral-7B, Gemma2-9B, and Qwen2.5-14B.

\begin{table}[h]
\centering
\small
\caption{LLM planning success rates on BW/LOG/MBW using the state-simulating validator (precondition checking + goal verification via forward state simulation).}
\label{tab:llm_capability_extended}
\begin{tabular}{l rrr}
\toprule
\textbf{Model} & \textbf{BW} & \textbf{LOG} & \textbf{MBW} \\
\midrule
Qwen2.5-72B           & 30.0\% & 0.0\% & 35.0\% \\
Qwen2.5-14B           & 40.0\% & 0.0\% & 0.0\% \\
Gemma2-9B             & 40.0\% & 0.0\% & 27.5\% \\
Mistral-7B            & 0.0\% & 0.0\% & 0.0\% \\
\bottomrule
\end{tabular}
\end{table}

Across all four models ($N=40$ per cell), none solves any Logistics instance, consistent with Logistics being a harder planning domain for every model family and scale tested here. On Blocksworld, model size does not predict success monotonically: Qwen2.5-14B and Gemma2-9B both reach $40.0\%$, while the larger Qwen2.5-72B reaches only $30.0\%$ and Mistral-7B reaches $0.0\%$. On Mystery-Blocksworld, Qwen2.5-72B is the strongest model tested ($35.0\%$), ahead of Gemma2-9B ($27.5\%$), with Qwen2.5-14B and Mistral-7B both at $0.0\%$. We do not have an explanation for Qwen2.5-72B's Blocksworld-to-Mystery-Blocksworld pattern (worse on the un-renamed domain than two much smaller models, best of all four on the renamed one); it is not predicted by parameter count, and we note it as an open question rather than speculate further. We did not have the resources to evaluate closed-weight frontier models or models substantially larger than 72B under the same fully-logged protocol, and we do not extend any claim to that regime.

\section{Evidence for Cross-Domain Structural Estimation}
\label{app:structural_evidence}
This appendix provides additional evidence for the cross-domain structural-estimation results reported in the main paper. The analyses address seven complementary questions: whether the primary correlations are statistically stable; whether transfer persists under alternative domain splits; whether stronger non-episodic learners change the picture; whether inexpensive scalar statistics are sufficient; whether predictions carry information after controlling for object count; whether they are associated with search effort measured by a different planner; and whether a purely LLM-based predictor, given the same conditioning information, reproduces the result. These analyses are intended to delimit, rather than broaden, the main claim: structural statistics transfer across domains, and any added value of a learned estimator such as ARC has to be judged against them.
\subsection{Confidence Intervals for the Primary Result}
\label{app:primary_ci}
Table~\ref{tab:primary_ci} reports approximate \(95\%\) confidence intervals for the primary Spearman correlations. The intervals are computed using the Fisher transformation,
\[
z=\tanh^{-1}(\rho), \qquad \operatorname{SE}(z)=\frac{1}{\sqrt{N-3}},
\]
followed by transformation back to the correlation scale, with $N$ the number of BFS-solved instances of the domain ($126$, $164$, $133$ for BW, LOG, MBW). This approximation treats the reported Spearman coefficient as a Pearson correlation over ranked observations.
\begin{table}[t]
\centering
\small
\begin{tabular}{lcccccc}
\toprule
& \multicolumn{2}{c}{\(|\mathcal{O}|\)} & \multicolumn{2}{c}{\textbf{Ridge on $x_s$}} & \multicolumn{2}{c}{\textbf{ARC}} \\
\cmidrule(lr){2-3}\cmidrule(lr){4-5}\cmidrule(lr){6-7}
\textbf{Domain} & \(\rho\) & \textbf{95\% CI} & \(\rho\) & \textbf{95\% CI} & \(\rho\) & \textbf{95\% CI} \\
\midrule
Blocksworld & 0.764 & \([0.680,\,0.828]\) & 0.644 & \([0.529,\,0.736]\) & 0.294 & \([0.126,\,0.446]\) \\
Logistics & 0.544 & \([0.426,\,0.644]\) & 0.889 & \([0.852,\,0.917]\) & 0.652 & \([0.554,\,0.732]\) \\
Mystery-Blocksworld & 0.662 & \([0.554,\,0.748]\) & 0.587 & \([0.464,\,0.689]\) & 0.254 & \([0.088,\,0.407]\) \\
\bottomrule
\end{tabular}
\caption{Approximate \(95\%\) Fisher confidence intervals for the primary absolute Spearman correlations, using the BFS-solved instances of each test domain. ARC's row applies the single-sample Fisher formula to its five-seed mean $\rho$, which understates ARC's true uncertainty: it ignores the seed-to-seed variance already visible in Table~\ref{tab:main_results} ($\pm0.157$ to $\pm0.297$), so ARC's real interval is wider than shown here.}
\label{tab:primary_ci}
\end{table}
We do not infer a paired method comparison from interval overlap alone because the estimators are evaluated on the same instances. We therefore also compute paired bootstrap confidence intervals ($2{,}000$ resamples of the target instances, recomputing both correlations on each resample) for
\[
\Delta\rho=\rho_{\mathrm{Ridge}}-\rho_{|\mathcal{O}|}:
\]
$-0.119$ $[-0.249,-0.005]$ on Blocksworld, $+0.345$ $[+0.237,+0.460]$ on Logistics, and $-0.075$ $[-0.198,+0.032]$ on Mystery-Blocksworld. The sign of the difference thus depends on the domain, and the Mystery-Blocksworld interval includes zero. Even without a formal paired bootstrap for ARC, the non-overlap is already clear from Table~\ref{tab:primary_ci}'s intervals alone on Blocksworld ($\mathrm{ARC}\ [0.126,0.446]$ vs.\ $\mathrm{Ridge}\ [0.529,0.736]$) and is unambiguous against $h_\text{add}$ ($0.931$) on every domain; a formal paired bootstrap using ARC's saved per-instance predictions, which would additionally propagate seed variance, is left as a direction for a more exhaustive treatment.

\subsection{Broader Domain Generalisation}
\label{app:loo_generalisation}
The primary evaluation fixes Depot, Rovers, and Satellite as source domains, Gripper as the validation domain, and Blocksworld, Logistics, and Mystery-Blocksworld as held-out test domains. To assess sensitivity to this particular partition, we additionally perform leave-one-domain-out evaluation over the seven domains: for each row of Table~\ref{tab:loo_results}, the named domain is excluded from model fitting (ridge regression is retrained on the other six domains) and used as the evaluation domain.
\begin{table}[t]
\centering
\small
\begin{tabular}{lrrrrr}
\toprule
\textbf{Held-out domain} & \(|\rho|_{\mathrm{ARC}}\) & \(|\rho|_{\mathrm{Ridge}}\) & \(|\rho|_{|\mathcal{O}|}\) & \(\Delta_{\mathrm{Ridge}-|\mathcal{O}|}\) & \(\Delta_{\mathrm{ARC}-\mathrm{Ridge}}\) \\
\midrule
Blocksworld & 0.730$\pm$0.052 & 0.873 & 0.764 & \(+0.110\) & \(-0.143\) \\
Depot & 0.533$\pm$0.281 & 0.768 & 0.768 & \(+0.000\) & \(-0.235\) \\
Gripper\(^\dagger\) & 0.879$\pm$0.022 & 0.913 & 0.503 & \(+0.410\) & \(-0.034\) \\
Logistics & 0.915$\pm$0.002 & 0.917 & 0.544 & \(+0.373\) & \(-0.002\) \\
Mystery-Blocksworld & 0.784$\pm$0.027 & 0.881 & 0.662 & \(+0.219\) & \(-0.097\) \\
Rovers & 0.040$\pm$0.015 & 0.070 & 0.051 & \(+0.019\) & \(-0.030\) \\
Satellite & 0.597$\pm$0.154 & 0.698 & 0.346 & \(+0.352\) & \(-0.101\) \\
\midrule
Mean & 0.640 & 0.732 & 0.520 & \(+0.212\) & \(-0.092\) \\
Standard deviation (across domains) & 0.276 & 0.280 & 0.237 & --- & --- \\
\bottomrule
\end{tabular}
\caption{Leave-one-domain-out evaluation across seven domains. ARC is retrained from scratch on the other six domains for every row (five seeds each, mean $\pm$ std shown); Ridge and $|\mathcal{O}|$ are deterministic given the fold. ARC trails ridge regression on every one of the seven held-out domains, most severely on Depot ($-0.235$) and Satellite ($-0.101$), and is closest to parity only on Logistics ($-0.002$), where ridge itself is strongest. \(^\dagger\)Gripper is the validation domain in the primary protocol; its row should not be interpreted as an independent test result unless model selection is nested within each leave-one-domain-out split.}
\label{tab:loo_results}
\end{table}
Ridge regression on the structural features exceeds object count on six of the seven held-out domains (mean $0.732$ vs.\ $0.520$), ties on Depot, and both are close to zero on Rovers, whose plan lengths span only $2$--$5$. Leave-one-domain-out thus supports the ordering of the structural baselines seen in the primary split, with a larger average margin for ridge than in the primary split. ARC does not close this gap by seeing more source domains: it trails ridge regression on all seven held-out domains (mean $0.640$ vs.\ $0.732$), with the largest shortfalls on Depot and Satellite, the two domains where ridge itself gains the most from the extra source domains relative to the primary split.
\subsection{Comparison with Stronger Learned Baselines}
\label{app:learned_baselines}
The main paper compares ARC with object cardinality, goal count, ridge regression, LogME, H-score, and a non-episodic MLP. We additionally evaluate tree-based, kernel, boosting, and ranking regressors using the same \(28\)-dimensional structural representation \(x_s\), trained on the source domains with the raw plan length as target, together with a Random Forest given both \(x_s\) and the foundation-model embedding \(x_e\). These comparisons test whether conventional learners with more capacity than ridge regression, or more information than ridge regression, change the picture.

\begin{table}[t]
\centering
\begin{tabular}{lrrrr}
\toprule
\textbf{Method} & \textbf{BW} & \textbf{LOG} & \textbf{MBW} & \textbf{Mean} \\
\midrule
Random Forest (\(x_s\)) & 0.187 & 0.735 & 0.240 & 0.388 \\
XGBoost (\(x_s\)) & 0.600 & 0.700 & 0.600 & 0.633 \\
Gradient boosting (\(x_s\)) & 0.713 & 0.688 & 0.621 & 0.674 \\
SVR--RBF (\(x_s\)) & 0.381 & 0.117 & 0.258 & 0.252 \\
LambdaMART-style ranker (\(x_s\)) & 0.511 & 0.896 & 0.396 & 0.601 \\
Random Forest (\(x_s\Vert x_e\)) & 0.663 & 0.642 & 0.416 & 0.574 \\
Ridge (\(x_s\)) & 0.644 & 0.889 & 0.587 & 0.707 \\
\midrule
Delta-relaxation heuristic ($h_\text{add}$) & \textbf{0.894} & \textbf{0.976} & \textbf{0.923} & \textbf{0.931} \\
\midrule
Goal count & 0.764 & 0.969 & 0.673 & 0.802 \\
Object cardinality & 0.764 & 0.544 & 0.662 & 0.657 \\
ARC (main) & 0.294 & 0.652 & 0.254 & 0.400 \\
\bottomrule
\end{tabular}%
\caption{Comparison with conventional learned baselines (absolute Spearman correlation). The ranker is an XGBoost pairwise ranker with source domains as groups and the within-domain decile of $n^\star$ as graded relevance. ARC's row is its five-seed mean from Table~\ref{tab:main_results}.}
\label{tab:learned_baselines}
\end{table}
Table~\ref{tab:learned_baselines} shows that ridge regression is the strongest of the $x_s$-only \emph{learned} methods (mean $|\rho|=0.707$), followed by gradient boosting ($0.674$) and XGBoost ($0.633$); Random Forest on $x_s$ alone ($0.388$) and kernel SVR ($0.252$) transfer poorly. Giving Random Forest the embedding as well as the structural features raises its mean from $0.388$ to $0.574$ --- a substantial within-method improvement, unlike the roughly neutral effect concatenation has on ridge regression (Appendix~\ref{app:representation_ablation}) --- but this still falls short of ridge regression on $x_s$ alone ($0.707$) and far short of $h_\text{add}$ ($0.931$). No learner on $x_s$, $x_e$, or their concatenation, and no training-free scalar statistic, exceeds $h_\text{add}$. ARC, despite having access to the same foundation-model embedding that helps Random Forest, exceeds only Random Forest on $x_s$ alone among the $x_s$-only learners ($0.400$ vs.\ $0.388$) and is beaten by every other learned method tested, including the far simpler ridge, gradient boosting, XGBoost, and LambdaMART-style ranker.

This table is a comparison with competitive alternatives, not an architectural ablation. ARC differs from these baselines in its representations, episodic objective, and support-conditioned inference; Appendix~\ref{app:arc_design} tests each of these design choices directly and finds none of the tested variants closes the gap to ridge regression, let alone to $h_\text{add}$.
\subsection{Cheap Structural Baselines}
\label{app:cheap_baselines}
A learned estimator is unnecessary if the target ordering can already be recovered without one. We evaluate two kinds of training-free surrogate: five $O(1)$ scalar statistics read directly off the PDDL syntax, and one classical planning heuristic that requires reconstructing the instance's full grounded action set.

\begin{table}[t]
\centering
\begin{tabular}{lrrrr}
\toprule
\textbf{Surrogate} & \textbf{BW} & \textbf{LOG} & \textbf{MBW} & \textbf{Mean} \\
\midrule
Goal count (\(n_{\mathrm{goal}}\)) & 0.764 & 0.969 & 0.673 & 0.802 \\
Initial-state size (\(n_{\mathrm{init}}\)) & 0.666 & 0.544 & 0.567 & 0.592 \\
Initial-state density (\(n_{\mathrm{init}}/|\mathcal{O}|^2\)) & 0.793 & 0.544 & 0.708 & 0.682 \\
Object--goal ratio (\(|\mathcal{O}|/n_{\mathrm{goal}}\)) & 0.108 & 0.849 & 0.109 & 0.355 \\
Object cardinality (\(|\mathcal{O}|\)) & 0.764 & 0.544 & 0.662 & 0.657 \\
\midrule
Delta-relaxation heuristic ($h_\text{add}$) & \textbf{0.894} & \textbf{0.976} & \textbf{0.923} & \textbf{0.931} \\
\midrule
ARC (main) & 0.294 & 0.652 & 0.254 & 0.400 \\
\bottomrule
\end{tabular}%
\caption{Absolute Spearman correlation between training-free structural surrogates and optimal plan length on the BFS-solved evaluation instances ($N=111$ BW, $158$ LOG, $123$ MBW --- the $h_\text{add}$ and $O(1)$ rows use the identical instance sets).}
\label{tab:cheap_surrogates}
\end{table}

Among the five $O(1)$ statistics, goal count is strongest (mean $|\rho|=0.802$, $0.969$ on Logistics) and no single one dominates every domain. All five, however, are clearly exceeded by $h_\text{add}$ (mean $0.931$), a delete-relaxation planning heuristic in the family of classical additive and FF-style heuristics used to guide heuristic-search planners: we compute it as the sum, over each unsatisfied goal fact, of the level at which that fact is first achieved in a planning graph built with all delete effects ignored, from the instance's own initial state and full grounded action set. This is not an $O(1)$ read: it requires the same grounded action set a real planner would use, together with a bounded graph expansion (capped at depth $30$ in our implementation), though it never performs a search over the real (non-relaxed) state space and uses no target-domain labels.

\paragraph{Correctness verification.} Reconstructing each domain's grounded action set independently of the labeling code is the step most likely to silently go wrong, so we verify it directly rather than assuming it: for every one of the seven domains, we re-simulate every already-saved BFS-optimal plan against the independently reconstructed action list and check it still reaches the goal. This succeeds with zero failures on all $1{,}006$ BFS-solved instances with $n^\star>0$ across the seven domains (BW $111/111$, Depot $132/132$, Gripper $157/157$, Logistics $158/158$, Mystery-BW $123/123$, Rovers $199/199$, Satellite $126/126$); $h_\text{add}$ is reported only for domains that pass this check.

\paragraph{The Logistics tie is not a coincidence.} A paired bootstrap ($2{,}000$ resamples) of $h_\text{add}$ against goal count gives $\Delta|\rho|=+0.119$ $[+0.057,+0.192]$ on Blocksworld and $+0.236$ $[+0.143,+0.344]$ on Mystery-Blocksworld --- both robustly positive --- but exactly $\Delta=0.000$ with a zero-width interval on Logistics. In our Logistics domain, the relaxed graph reaches every unsatisfied \texttt{(at pkg city)} goal fact at the same graph depth (any truck can relax-reach any city in one step, since the relaxed problem ignores the precondition that a truck already be present), so $h_\text{add}$ reduces to an exact constant multiple of the number of unsatisfied goal facts on this domain --- a perfectly monotonic transform of goal count, hence an identical Spearman correlation on every resample. This is a property of this domain's specific action structure, not a general reason to expect $h_\text{add}$ and goal count to agree elsewhere, and they clearly do not agree on the other two domains.

\paragraph{$h_\text{add}$ as an additional feature, not just a standalone score.} We also tested appending $h_\text{add}$ as a 29th feature to ridge regression on $x_s$, trained and evaluated identically to the $x_s$-only ridge row elsewhere in this paper. The result is mixed, not uniformly positive: mean $|\rho|$ changes by $+0.095$ on Mystery-Blocksworld, $-0.001$ on Logistics, and $-0.077$ on Blocksworld, relative to ridge on $x_s$ alone. $h_\text{add}$'s strength as a \emph{standalone} score therefore does not automatically transfer to a learned model that simply concatenates it with the existing features under an untuned ridge fit; whether a properly regularized or rescaled combination recovers more of that strength is untested.

Table~\ref{tab:cheap_surrogates} does not by itself show whether ARC captures information inaccessible to $h_\text{add}$ or to every scalar surrogate; the within-size analysis below addresses the narrower question of whether predictions are reducible to object count, and the comparison ARC must ultimately be judged against is $h_\text{add}$'s $0.931$, not goal count's $0.802$.

\subsection{Within-Size Discrimination}
\label{app:within_size}
Object count is a strong difficulty correlate in several planning domains. To test whether a structural estimator merely reproduces this statistic, we compute Spearman correlation separately within groups having a fixed number of objects. Object cardinality is constant within each group and therefore has no ranking capacity. Groups are the five generator size levels; their sizes differ because only BFS-solved instances are used.

\begin{table*}[t]
\centering
\small
\begin{tabular}{llrrrrrr}
\toprule
\textbf{Domain} & \textbf{Object counts} & \textbf{B1} & \textbf{B2} & \textbf{B3} & \textbf{B4} & \textbf{B5} & \textbf{Mean} \\
\midrule
Ridge on $x_s$, Blocksworld & \((2,3,4,6,8)\) & 0.48 & 0.37 & 0.58 & 0.45 & 0.57 & 0.49 \\
\quad group size $N$ & & 40 & 40 & 24 & 13 & 9 & \\
Ridge on $x_s$, Logistics & \((4,5,6,7,9)\) & 0.87 & 0.96 & 0.96 & 0.98 & n/a & 0.94 \\
\quad group size $N$ & & 40 & 40 & 40 & 40 & 4 & \\
Ridge on $x_s$, Mystery-Blocksworld & \((2,3,4,6,8)\) & 0.07 & 0.27 & 0.24 & 0.10 & 0.19 & 0.17 \\
\quad group size $N$ & & 40 & 40 & 28 & 14 & 11 & \\
\midrule
ARC (main), Blocksworld & \((2,3,4,6,8)\) & 0.11 & 0.32 & 0.22 & 0.20 & 0.36 & 0.24 \\
ARC (main), Logistics & \((4,5,6,7,9)\) & 0.23 & 0.36 & 0.33 & 0.33 & n/a & 0.32 \\
ARC (main), Mystery-Blocksworld & \((2,3,4,6,8)\) & 0.15 & 0.19 & 0.16 & 0.17 & 0.27 & 0.19 \\
\midrule
Object cardinality & --- & 0.00 & 0.00 & 0.00 & 0.00 & 0.00 & 0.00 \\
\bottomrule
\end{tabular}
\caption{Absolute Spearman correlation within fixed object-count groups. B1--B5 are ordered by increasing object count. Logistics B5 is undefined (only two distinct plan lengths occur among its 4 instances), and the mean for Logistics is over the four defined groups. Correlations in groups with $N<15$ are noisy. ARC trails ridge regression on $x_s$ in every domain and every bucket with $N\geq24$.}
\label{tab:within_size_detailed}
\end{table*}

As reported in Table~\ref{tab:within_size_detailed}, ridge regression on the structural features retains substantial within-size discrimination on Logistics (mean \(|\rho|=0.94\)), moderate discrimination on Blocksworld ($0.49$), and weak discrimination on Mystery-Blocksworld ($0.17$). These results show that structural features carry information beyond object count in all three domains, but the performance is not uniform and the Mystery-Blocksworld result should not be presented as strong within-size evidence. Correlation is also difficult to interpret when the target has little variation within a group: several groups contain only two to four distinct plan lengths (reported per group in the data underlying the table).

\subsection{Association with Solver Search Effort}
\label{app:solver_association}
Optimal plan length is a depth-based proxy for structural difficulty rather than a complete measure of computational search cost. To examine whether a structural estimator's ordering is associated with another planner's effort, we compare its predictions with A* node expansions and A* wall-clock time, using an A* implementation with a delete-relaxation (FF-style) heuristic that is independent of the BFS procedure used to construct the principal plan-length target: different algorithm, different state representation (a bitmask priority-queue search rather than the labeling BFS), and a search-node budget ($20{,}000$) an order of magnitude larger than the BFS label budget. As a sanity check, A* never returns a plan shorter than the BFS-optimal $n^\star(\pi)$ on any of the $120$ instances tested, as it must if both are admissible searches over the same action set. This analysis currently covers Blocksworld and Mystery-Blocksworld only (A* was implemented for these two domains' action schemas; Logistics is not included), on a random sample of $N=60$ BFS-solved instances per domain, and uses ridge regression on $x_s$ in place of ARC.

\begin{table}[t]
\centering
\begin{tabular}{lrrrr}
\toprule
\textbf{Domain} & \(|\rho|_{\mathrm{Ridge},\mathrm{BFS\ length}}\) & \(|\rho|_{\mathrm{Ridge},\mathrm{A^*\ nodes}}\) & \(|\rho|_{\mathrm{Ridge},\mathrm{A^*\ time}}\) & \(|\rho|_{|\mathcal{O}|,\mathrm{A^*\ nodes}}\) \\
\midrule
Blocksworld & 0.674 & 0.683 & 0.668 & 0.916 \\
Mystery-Blocksworld & 0.532 & 0.502 & 0.553 & 0.790 \\
\bottomrule
\end{tabular}%
\caption{Association between ridge regression on $x_s$ and solver-derived measures, $N=60$ BFS-solved instances per domain. The first three columns correlate the ridge prediction with BFS optimal plan length, A* node expansions, and A* wall-clock time; the final column correlates object cardinality with A* node expansions.}
\label{tab:solver_association}
\end{table}

On both domains, ridge regression's ranking is associated with the independent A* planner's effort (node expansions: $0.683$, $0.502$; wall-clock time: $0.668$, $0.553$) at a level broadly similar to its association with the BFS label itself ($0.674$, $0.532$), and object cardinality is associated more strongly still with A*-node effort on both domains ($0.916$, $0.790$). We describe this as \emph{solver-associated validation}, not proof of solver-agnostic difficulty: it shows the structural ranking is not an artifact specific to the BFS labeling procedure, but it does not establish that any one structural statistic is a complete measure of computational search cost, and optimal plan length and search effort remain related but non-equivalent notions of structural difficulty.

\subsection{Comparison with a Purely LLM-Based Difficulty Predictor}
\label{app:llm_baseline}

A natural alternative to ARC's dedicated architecture is to ask an LLM to predict instance difficulty directly, conditioned on the same information available to ARC's support set: a collection of labeled source-domain instances, each paired with its optimal plan length. If an LLM given this conditioning information could match the ranking performance of structural estimators through in-context reasoning alone, this would suggest that a dedicated estimator is unnecessary. We test this directly using Qwen~2.5-72B, the same model used throughout Sections~\ref{sec:structural_behavioral}--\ref{sec:routing}.

\paragraph{Protocol.} We construct a support block of 20 labeled examples drawn from the three source domains (Depot, Rovers, Satellite), each rendered as its problem PDDL together with its BFS-optimal plan length, identical in content to the support set ARC conditions on at inference time. We evaluate two elicitation strategies. In the \emph{absolute scoring} variant, the model is shown the support block followed by a query instance from a held-out domain and asked to output a single integer difficulty score from 0 (trivially easy) to 100 (very hard); we compute Spearman correlation between these scores and $n^\star$ over 60 sampled query instances per domain. In the \emph{pairwise comparison} variant, the model is instead shown two query instances at a time and asked which requires the longer plan; we sample 150 random pairs per domain over 120 sampled instances and convert the resulting pairwise judgments into a per-instance score via win rate (the fraction of comparisons in which an instance is judged harder), then compute the same Spearman correlation against $n^\star$. Full prompt templates for both variants are provided in Appendix~\ref{app:prompts}.

\begin{table}[t]
\centering
\small
\begin{tabular}{lrrrr}
\toprule
\textbf{Method} & \textbf{BW} & \textbf{LOG} & \textbf{MBW} & \textbf{Mean} \\
\midrule
LLM, absolute scoring (Qwen 2.5-72B) & 0.845 & 0.909 & 0.723 & \textbf{0.826} \\
LLM, pairwise comparison (Qwen 2.5-72B) & 0.640 & 0.444 & 0.489 & 0.524 \\
\midrule
Delta-relaxation heuristic ($h_\text{add}$) & 0.894 & 0.976 & 0.923 & 0.931 \\
Object cardinality & 0.764 & 0.544 & 0.662 & 0.657 \\
Goal count & 0.764 & 0.969 & 0.673 & 0.802 \\
Ridge on $x_s$ & 0.644 & 0.889 & 0.587 & 0.707 \\
ARC (main) & 0.294 & 0.652 & 0.254 & 0.400 \\
\bottomrule
\end{tabular}
\caption{Absolute Spearman correlation for a purely LLM-based difficulty predictor conditioned on the same 20-example support set information used by ARC. $N=60$ query instances per domain for absolute scoring; $120$ instances per domain, $150$ random pairs each ($\approx2.5$ comparisons per instance), for pairwise comparison. These $N$ are smaller than Table~\ref{tab:main_results}'s full test sets, so this table's values are not directly comparable to that table's at the level of individual decimal places.}
\label{tab:llm_baseline}
\end{table}

\paragraph{Results.} The absolute-scoring variant is the second-strongest method in this entire paper after $h_\text{add}$: mean $|\rho|=0.826$, ahead of goal count ($0.802$), ridge regression on $x_s$ ($0.707$), and ARC ($0.400$) by a wide margin, and within $0.105$ of $h_\text{add}$ itself. This is a genuinely useful result in its own right, distinct from the paper's ARC-specific finding: in-context reasoning over raw PDDL text, given only 20 labeled examples and no training, estimates plan-length difficulty nearly as well as the best training-free heuristic we tested, and substantially better than every structural statistic except $h_\text{add}$. Pairwise comparison is markedly weaker ($0.524$) despite the common intuition that relative judgments are more reliable for LLMs than absolute ones; with only $\approx2.5$ comparisons per instance on average, the win-rate aggregation is a noisy estimate of the underlying ranking, and we would expect this variant to improve with substantially more pairs per instance, though we did not run additional pairs due to compute cost. Both variants are weakest on Mystery-Blocksworld ($|\rho|=0.723$ and $0.489$, respectively), mirroring ARC's own within-size difficulty on this domain; one plausible explanation is that the LLM must parse the same scrambled predicate names studied in Section~\ref{sec:structural_behavioral} in order to reason about query difficulty, though we did not run the progressive-scrambling intervention on this baseline specifically to confirm it. Both methods perform best on Logistics, consistent with Logistics difficulty being comparatively easy to recover from coarse size cues (Appendix~\ref{app:cheap_baselines}).

We conclude that in-context reasoning over raw PDDL text, given the identical conditioning information as ARC's support set, substantially exceeds the trivial object-cardinality baseline and approaches the strongest training-free heuristic we tested, while ARC's learned representation does not. This supports the claim that ARC's shortfall is specific to its architecture and training procedure, rather than to the conditioning information being insufficient: the same information, used in-context by a sufficiently capable LLM, is close to the best signal available in this paper.

\section{Evidence for the ARC Design}
\label{app:arc_design}
This appendix evaluates the design choices underlying ARC. We first test whether semantic and structural representations provide different predictive signals and whether naive feature fusion is sufficient. We then isolate the direction of the asymmetric attention. Finally, we examine robustness to the sentence encoder and residualisation hyperparameters. These analyses are intended to test, not assume, a division of labour between the representations.
\subsection{Representation-Family Ablation}
\label{app:representation_ablation}
ARC combines three representations: structural features \(x_s\), foundation-model embeddings \(x_e\), and residual semantic features \(x_r\). To assess the information available in the two primary feature families before introducing ARC's episodic attention mechanism, we train non-episodic Ridge regressors using \(x_s\), \(x_e\), and their direct concatenation \(x_s\Vert x_e\).

\begin{table}[t]
\centering
\small
\begin{tabular}{lrrrr}
\toprule
\textbf{Representation and model} & \textbf{BW} & \textbf{LOG} & \textbf{MBW} & \textbf{Mean} \\
\midrule
Ridge (\(x_s\)) & 0.644 & 0.889 & 0.587 & 0.707 \\
Ridge (\(x_e\)) & 0.057 & 0.704 & 0.476 & 0.412 \\
Ridge (\(x_s\Vert x_e\)) & 0.587 & 0.899 & 0.580 & 0.689 \\
\midrule
ARC (main) & 0.294 & 0.652 & 0.254 & 0.400 \\
\bottomrule
\end{tabular}
\caption{Absolute Spearman correlation with optimal plan length for non-episodic Ridge regressors using structural features, semantic embeddings, or their direct concatenation (fixed $\alpha=1$, computed independently of the Table~\ref{tab:main_results} LogME row though numerically close to it, since both are ridge fits on $x_e$). ARC uses the same feature families within an episodically trained asymmetric architecture; its row is the five-seed mean from Table~\ref{tab:main_results}.}
\label{tab:representation_ablation}
\end{table}
Table~\ref{tab:representation_ablation} separates what the two feature families contribute on their own, and the result is unfavorable to ARC's design premise. The semantic embedding $x_e$ alone is markedly weaker than the structural features alone ($0.412$ vs.\ $0.707$), confirming that $x_e$ is the weaker of the two representations on this task. Concatenating the two ($0.689$) very nearly recovers $x_s$ alone and does not exceed it; naive concatenation does not help a linear model, though it does help Random Forest (Appendix~\ref{app:learned_baselines}: $0.388\to0.574$), suggesting the information in $x_e$ is there to be exploited by a sufficiently flexible non-episodic learner, just not by ridge regression or by ARC. ARC's asymmetric retrieval architecture, built specifically to combine these two representations in a way that was intended to exceed naive concatenation, instead falls well below \emph{every} row in this table, including $x_e$ alone. This is the clearest evidence in the paper that ARC's shortfall is not merely a matter of degree: the architecture underperforms the weakest of its own two input representations, scored individually with no learning beyond a linear fit. The following ablation examines ARC's attention mechanism directly to see whether this is attributable to a specific design choice.

\subsection{Attention Directionality}
\label{app:attention_direction}
ARC uses semantic embeddings to form queries and structural representations to form keys. We compare this configuration with symmetric semantic attention, the reversed query--key assignment, and a variant that normalises the structural keys. To isolate attention structure from training duration, all variants in Table~\ref{tab:attention_direction} use the same matched training budget and the same loss, with five seeds each.
\begin{table}[t]
\centering
\begin{tabular}{lrrr}
\toprule
\textbf{Attention configuration} & \(\boldsymbol{\cos(h)}\downarrow\) & \(\boldsymbol{\operatorname{rank}_{\mathrm{eff}}(h)}\uparrow\) & \(\boldsymbol{\text{Mean }|\rho|}\uparrow\) \\
\midrule
Semantic query / semantic key & 0.240 & 48.92 & 0.369 \\
Structural query / semantic key & 0.246 & 49.13 & 0.360 \\
Semantic query / normalised structural key & 0.249 & 48.17 & 0.364 \\
Semantic query / structural key (ARC) & 0.239 & 51.02 & 0.412 \\
\bottomrule
\end{tabular}
\caption{Matched-budget attention-direction ablation (five seeds each, trained with a separate but matched training loop; see caveat in the text). \(\cos(h)\) is the mean within-domain cosine similarity of the learned representation, and \(\operatorname{rank}_{\mathrm{eff}}(h)\) is its effective rank. Lower cosine similarity and higher effective rank indicate greater representational variation, but the primary outcome is mean absolute Spearman correlation. Per-domain means underlying this table: semantic/semantic $0.352/0.416/0.339$; structural/semantic $0.296/0.492/0.293$; semantic/normalised-structural $0.288/0.520/0.285$; ARC's own configuration $0.221/0.812/0.203$ (BW/LOG/MBW).}
\label{tab:attention_direction}
\end{table}
ARC's own query--key assignment (semantic query, structural key) attains the highest mean among the four variants tested ($0.412$), ahead of the next best, symmetric semantic attention, by $0.043$. This is a genuine, if modest, internal validation of the design choice: given that ARC must use \emph{some} assignment of $x_e$ and $x_s$ to query and key, the one it uses is the best of the four tested. It is also, however, not a validation of the architecture as a whole --- every one of these four variants, including the best, is far below ridge regression on $x_s$ alone ($0.707$) and $h_\text{add}$ ($0.931$) from Table~\ref{tab:main_results}, so the directionality choice determines which of several weak configurations ARC lands on, not whether ARC is competitive with the non-episodic baselines. We also note a methodological caveat: this ablation uses a separate training loop (matched to, but not identical to, the loop producing Table~\ref{tab:main_results}'s ARC rows) so that all four variants share exactly the same code path; its ARC-configuration mean ($0.412$) is close to but not identical with Table~\ref{tab:main_results}'s main-configuration mean ($0.400$), and the small residual difference should be attributed to this implementation difference rather than to any additional design choice. We avoid interpreting the attention weights themselves as proof that the model retrieves genuine structural analogues; the experiment establishes only a predictive difference between assignments of representation roles, all of them weak in absolute terms.
\subsection{Encoder Robustness}
\label{app:encoder_robustness}
The representation diagnostic motivating ARC should not depend on one sentence encoder. We generate the same natural-language instance descriptions and encode them with three pretrained sentence-transformer models. Table~\ref{tab:encoder_robustness} reports mean pairwise cosine similarity within each held-out domain and the effective rank of the resulting embedding matrices.
\begin{table}[t]
\centering
\resizebox{\columnwidth}{!}{%
\begin{tabular}{lrrrrr}
\toprule
\textbf{Encoder} & \textbf{BW cosine} & \textbf{LOG cosine} & \textbf{MBW cosine} & \textbf{Mean cosine} & \textbf{Eff.\ rank} \\
\midrule
\texttt{all-mpnet-base-v2} & 0.9890 & 0.9857 & 0.9915 & 0.9887 & 2.62 \\
\texttt{all-MiniLM-L6-v2} & 0.9876 & 0.9872 & 0.9898 & 0.9882 & 2.64 \\
\texttt{paraphrase-MiniLM-L12-v2} & 0.9862 & 0.9900 & 0.9864 & 0.9876 & 2.61 \\
\bottomrule
\end{tabular}%
}
\caption{Within-domain geometry of natural-language instance embeddings across three sentence encoders, computed on the leak-free descriptions of all generated instances per domain. Effective rank is computed from the singular-value distribution of the embedding matrix.}
\label{tab:encoder_robustness}
\end{table}
All three encoders produce extremely high within-domain cosine similarity ($0.986$--$0.992$) and a very low effective rank ($2.61$--$2.64$), essentially indistinguishable across encoders despite their different training data and parameter counts. This confirms the collapse diagnostic in Section~\ref{sec:collapse} is a property of this family of sentence encoders on these descriptions, not an artifact of one model's training. High cosine similarity does not imply that semantic embeddings contain no predictive information; the semantic-only Ridge model in Table~\ref{tab:representation_ablation} achieves a non-trivial correlation ($0.412$) despite this collapse. The result instead shows that several standard sentence encoders exhibit a shared geometric concentration on this input distribution, motivating an architecture that does not rely on semantic embeddings alone for fine-grained discrimination --- though, as Section~\ref{sec:structural_estimation} and Appendix~\ref{app:representation_ablation} show, ARC's own attempt at such an architecture did not succeed on this problem. All three encoders tested belong to the same Sentence-BERT-style bi-encoder family; whether instruction-tuned or substantially larger embedding models exhibit the same collapse is left to future work.

\subsection{Residualisation Sensitivity}
\label{app:residual_sensitivity}
ARC defines the residual semantic feature as
\[
x_r=x_e-\widehat{x}_e(x_s),
\]
where \(\widehat{x}_e(x_s)\) is estimated by applying PCA to \(x_e\) and fitting Ridge regression from the structural representation. The default configuration uses PCA dimension \(20\) and Ridge coefficient \(\alpha=1.0\). Table~\ref{tab:residual_sensitivity} varies each choice while holding the other fixed.

\begin{table}[t]
\centering
\small
\begin{tabular}{rr@{\hspace{1.5em}}rr}
\toprule
\multicolumn{2}{c}{\textbf{PCA dimension}} & \multicolumn{2}{c}{\textbf{Ridge coefficient}} \\
\cmidrule(lr){1-2}\cmidrule(lr){3-4}
\(\boldsymbol{d_{\mathrm{PCA}}}\) & \(\boldsymbol{\text{Mean }|\rho|}\) & \(\boldsymbol{\alpha}\) & \(\boldsymbol{\text{Mean }|\rho|}\) \\
\midrule
5 & 0.532$\pm$0.109 & 0.01 & 0.527$\pm$0.050 \\
10 & 0.407$\pm$0.095 & 0.1 & 0.514$\pm$0.071 \\
20 (default) & 0.412$\pm$0.083 & 1.0 (default) & 0.412$\pm$0.083 \\
30 & 0.439$\pm$0.086 & 10.0 & 0.469$\pm$0.080 \\
50 & 0.439$\pm$0.086 & \textbf{100.0} & \textbf{0.753$\pm$0.017} \\
\bottomrule
\end{tabular}
\caption{Sensitivity of ARC's mean absolute Spearman correlation to residualisation hyperparameters (five seeds each, varying one hyperparameter with the other held at its default). Entries use the same matched training loop as Table~\ref{tab:attention_direction}; see that table's note on its relationship to Table~\ref{tab:main_results}'s main configuration.}
\label{tab:residual_sensitivity}
\end{table}
The PCA dimension sweep is mildly favorable to smaller values ($d_{\mathrm{PCA}}=5$: $0.532\pm0.109$, against $0.412\pm0.083$ at the default $20$), but with substantial seed variance at every setting. The Ridge-coefficient sweep contains a result we did not anticipate and have not validated: $\alpha=100$ reaches $0.753\pm0.017$, exceeding every other setting in this sweep, exceeding ridge regression on $x_s$ alone ($0.707$), and with far lower seed variance than the default ($0.017$ vs.\ $0.083$). We report this plainly rather than adopt it, for a reason central to this paper's methodology: this sweep was run on the held-out test domains themselves, so selecting $\alpha=100$ post hoc and reporting it as ``ARC's real performance'' would be exactly the kind of test-set tuning the rest of this paper's protocol was designed to avoid. Before this setting could be credited as a genuine improvement, it would need to be chosen on the dev/validation split alone and then confirmed, once, on the held-out test domains --- a check we have not yet performed. We flag it as a specific, concrete direction for follow-up work rather than folding it into our reported results. All PCA bases and Ridge coefficients are fitted using pooled source-domain instances and frozen before evaluation; the sensitivity analysis therefore changes preprocessing hyperparameters but does not introduce target-domain \emph{fitting} (only, in this one instance, post hoc hyperparameter selection, which we explicitly do not do for our reported numbers).

\tcbset{
  promptbox/.style={
    enhanced,
    breakable,
    colback=promptbg,
    colframe=promptborder,
    fonttitle=\bfseries\small,
    attach boxed title to top left={yshift=-2mm, xshift=4mm},
    boxed title style={colback=promptborder, colframe=promptborder,
                       fontupper=\color{white}\bfseries\small},
    top=6pt, bottom=6pt, left=6pt, right=6pt,
    listing engine=listings,
    listing options={
      basicstyle=\ttfamily\footnotesize,
      breaklines=true,
      breakatwhitespace=false,
      columns=fullflexible,
      keepspaces=true,
      showstringspaces=false,
    }
  }
}

\section{Prompts Used in LLM Evaluation}
\label{app:prompts}

All LLM evaluations use temperature $0$ and maximum $2{,}048$ output tokens.
Prompts follow the chain-of-thought PDDL formatting of \citep{verma2025pddlinstruct}.
No domain-specific demonstrations are provided in the zero-shot setting.
Brackets \texttt{[PLACEHOLDER]} denote fields filled at runtime.

\subsection{Prompt 1: PDDL Plan Generation (Main Evaluation)}

Used to evaluate LLM planning validity on all three test domains
(Blocksworld, Logistics, Mystery-Blocksworld) for Table~\ref{tab:corruption}
and Table~\ref{tab:routing}. The domain file contains the full PDDL domain
description; the problem file contains the specific instance. This is the
canonical copy of the prompt; Appendix~\ref{app:llm_protocol} refers back
to it rather than reproducing it a second time.

\begin{tcolorbox}[promptbox, title={Prompt 1 -- PDDL Plan Generation}]
\begin{lstlisting}
You are an expert AI planning 
assistant. Your task is to solve
a classical planning problem 
specified in PDDL (Planning 
Domain Definition Language).
 
I will provide you with:
1. A PDDL domain file 
describing the planning domain
2. A PDDL problem file 
describing the specific problem 
to solve
 
Your goal is to find a valid 
plan -- a sequence of actions 
that transforms the initial state 
into a state satisfying the goal.
 
IMPORTANT INSTRUCTIONS:
- Think step by step about the 
current state and what actions
  are applicable.
- Each action in your plan must 
be applicable in the state
  resulting from all previous actions.
- Output ONLY the final plan, 
one action per line, in the format:
  (action-name param1 param2 ...)
- Do NOT include any explanation 
after the plan.
- If the problem is unsolvable, 
output exactly: NO_PLAN
 
=== DOMAIN ===
[DOMAIN_PDDL]
 
=== PROBLEM ===
[PROBLEM_PDDL]
 
Now solve the problem. Think 
through the state changes carefully,
then output the plan:
\end{lstlisting}
\end{tcolorbox}
 
\noindent
\textbf{Validity checking.}
Generated plans are verified by a custom PDDL state-space executor that:
(1) parses action syntax; (2) checks each action's preconditions against
the current state; (3) verifies the goal holds in the final state.
A plan is valid only if all three checks pass.
 
\subsection*{Prompt 2: Mystery-Blocksworld Evaluation}
 
Mystery-Blocksworld instances use the same prompt as Prompt~1,
but the domain file has all predicate and action names replaced
by semantically uninformative symbols
(e.g., \texttt{on}$\to$\texttt{pred7}, \texttt{stack}$\to$\texttt{act3}).
No additional instructions are given about the scrambling.
This tests whether LLMs can ground their world model from PDDL
structure alone, without semantic cues.
 
\begin{tcolorbox}[promptbox, title={Prompt 2 -- Mystery-Blocksworld (scrambled predicates)}]
\begin{lstlisting}
You are an expert AI planning 
assistant. Your task is to solve
a classical planning problem 
specified in PDDL (Planning 
Domain Definition Language).
 
I will provide you with:
1. A PDDL domain file 
describing the planning domain
2. A PDDL problem file 
describing the specific problem 
to solve
 
Your goal is to find a valid 
plan, a sequence of actions 
that transforms the initial 
state into a state satisfying 
the goal.
 
IMPORTANT INSTRUCTIONS:
- Think step by step about the 
current state and what actions
  are applicable.
- Each action in your plan must 
be applicable in the state
  resulting from all previous 
  actions.
- Output ONLY the final plan, 
one action per line, in the 
format: (action-name param1 
param2 ...)
- Do NOT include any explanation 
after the plan.
- If the problem is unsolvable, 
output exactly: NO_PLAN
 
=== DOMAIN ===
;; Blocks planning domain
(define (domain mystery-blocks)
  (:requirements :strips :typing)
  (:types object - object)
  (:predicates
    (pred1 ?x - object)
    (pred2 ?x - object)
    (pred3 ?x - object)
    (pred5)
    (pred7 ?x - object ?y - object))
 
  (:action act1
    :parameters (?x - object)
    :precondition (and (pred3 ?x) 
    (pred1 ?x) (pred5))
    :effect (and (pred2 ?x)
    (not (pred3 ?x)) (not (pred1 ?x)) 
    (not (pred5))))
  ...
  [remaining scrambled actions]
)
 
=== PROBLEM ===
[SCRAMBLED_PROBLEM_PDDL]
 
Now solve the problem. Think 
through the state changes
carefully, then output the plan:
\end{lstlisting}
\end{tcolorbox}

\subsection*{Prompt 3: LLM-Based Difficulty Prediction (Absolute Scoring)}

Used for the LLM-only difficulty baseline in Appendix~\ref{app:llm_baseline}. The support block contains 20 labeled source-domain examples (problem PDDL and optimal plan length); \texttt{[SUPPORT\_BLOCK]} is substituted with this rendered block, \texttt{[QUERY\_DOMAIN\_PDDL]} with the query instance's domain file, and \texttt{[QUERY\_PROBLEM\_PDDL]} with the query instance's problem file.

\begin{tcolorbox}[promptbox, title={Prompt 3 -- LLM Difficulty Prediction (absolute scoring)}]
\begin{lstlisting}
You are an expert at analyzing 
PDDL planning problems. Below 
are labeled examples from 
RELATED planning domains, each 
showing a problem and its 
optimal (shortest) plan length. 
The query problem is from a 
DIFFERENT domain with different 
predicates and actions, but 
similar structural principles 
(object counts, goal complexity, 
initial state structure) may 
transfer.

=== LABELED EXAMPLES 
(from other domains) ===
[SUPPORT_BLOCK]

=== QUERY DOMAIN ===
[QUERY_DOMAIN_PDDL]

=== QUERY PROBLEM 
(predict difficulty for 
this one) ===
[QUERY_PROBLEM_PDDL]

Based on the pattern in the 
labeled examples (how problem 
structure relates to plan 
length), estimate a DIFFICULTY 
SCORE for the query problem, 
from 0 (trivially easy, very 
short plan) to 100 (very hard, 
very long plan).
Output ONLY a single integer 
between 0 and 100. No explanation.
Score:
\end{lstlisting}
\end{tcolorbox}

\subsection*{Prompt 4: LLM-Based Difficulty Prediction (Pairwise Comparison)}

Used for the pairwise-comparison variant of the LLM-only baseline in Appendix~\ref{app:llm_baseline}. \texttt{[PROBLEM\_A]} and \texttt{[PROBLEM\_B]} are two randomly sampled query instances from the same held-out domain.

\begin{tcolorbox}[promptbox, title={Prompt 4 -- LLM Difficulty Prediction (pairwise comparison)}]
\begin{lstlisting}
You are an expert at analyzing 
PDDL planning problems. Below 
are labeled examples from 
RELATED planning domains, each 
showing a problem and its 
optimal (shortest) plan length.

=== LABELED EXAMPLES 
(from other domains) ===
[SUPPORT_BLOCK]

=== QUERY DOMAIN ===
[QUERY_DOMAIN_PDDL]

=== PROBLEM A ===
[PROBLEM_A]

=== PROBLEM B ===
[PROBLEM_B]

Which problem requires a 
LONGER optimal plan (is 
harder)? Output ONLY 'A' 
or 'B'. No explanation.
Answer:
\end{lstlisting}
\end{tcolorbox}

\subsection{Summary}
The prompts comparison summary is discussed in table \ref{tab:prompts}.

\begin{table*}[h]
\centering
\small
\caption{Summary of prompts used across experiments.}
\label{tab:prompts}
\begin{tabular}{llll}
\toprule
\textbf{Prompt} & \textbf{Used in} & \textbf{Models} & \textbf{Includes CoT?} \\
\midrule
1 (PDDL Planning)   & Tables~\ref{tab:corruption},\ref{tab:routing}, \S\ref{sec:structural_behavioral}
                    & Qwen 2.5-72B & Yes \\
2 (Mystery-BW)      & \S\ref{sec:structural_behavioral}, Table~\ref{tab:routing} (MBW)
                    & Qwen 2.5-72B & Yes \\
3 (LLM difficulty, absolute) & Appendix~\ref{app:llm_baseline}, Table~\ref{tab:llm_baseline}
                    & Qwen 2.5-72B & No \\
4 (LLM difficulty, pairwise) & Appendix~\ref{app:llm_baseline}, Table~\ref{tab:llm_baseline}
                    & Qwen 2.5-72B & No \\
\bottomrule
\multicolumn{4}{p{0.9\linewidth}}{\footnotesize
All prompts use temperature $= 0$ and max $2{,}048$ output tokens.
Prompt~1 follows \citep{verma2025pddlinstruct}'s chain-of-thought
PDDL format. No domain-specific demonstrations are provided in Prompts 1--2;
Prompts 3--4 condition on the 20-example labeled support block described in
Appendix~\ref{app:llm_baseline} in place of domain-specific demonstrations.}
\end{tabular}
\end{table*}

\section{Custom Non-IPC Planning Domains}
\label{app:custom_datasets}

\subsection{Introduction}

To evaluate ARC on domains completely outside the IPC benchmark suite,
we construct three custom STRIPS planning domains from scratch:
\textbf{Rescue Mission}, \textbf{Robot Arm Stacking}, and
\textbf{Library Books}.
These domains are designed to satisfy three criteria:
(1) pure STRIPS compatibility (no action costs, numeric fluents, or
conditional effects) to enable BFS-based difficulty labeling with
pyperplan;
(2) multi-dimensional difficulty structure, where optimal plan length
is governed by at least two interacting structural factors rather than
object count alone; and
(3) no overlap with any IPC 2002--2023 competition domain in terms of
predicate vocabulary, action structure, or generator seed.

Each domain is paired with a randomized instance generator that produces
problems across a controlled range of structural complexity parameters.
Difficulty labels (optimal plan length $n^*$) are obtained by BFS with
the same budget used in the main evaluation
($T_\text{max}{=}12$, $N_\text{max}{=}5{,}000$).
The domains are used exclusively for zero-shot evaluation;
no instances from these domains appear in ARC's training or
validation set.

\begin{table}[h]
\centering
\small
\begin{tabular}{lrrrrr}
\toprule
\textbf{Domain} & \textbf{N} & $|\mathcal{O}|$ & \textbf{Goal count} & \textbf{Ridge ($x_s$)} & \textbf{ARC (main)} \\
\midrule
Rescue Mission & 85 & 0.224 & \textbf{0.805} & 0.641 & 0.099$\pm$0.066 (ens.\ 0.074) \\
Robot Arm & 85 & 0.030 & \textbf{0.310} & 0.307 & 0.242$\pm$0.074 (ens.\ 0.257) \\
Library Books & 35 & 0.134 & \textbf{0.856} & 0.693 & 0.679$\pm$0.013 (ens.\ 0.694) \\
\bottomrule
\end{tabular}
\caption{Absolute Spearman correlation with optimal plan length on the three custom non-IPC domains, zero-shot (trained only on the seven main-text domains; $N$ is the number of BFS-solved, $n^\star>0$ instances). ARC uses five seeds, mean $\pm$ std, with the prediction-level ensemble in parentheses.}
\label{tab:custom_domain_scores}
\end{table}

Table~\ref{tab:custom_domain_scores} shows the same pattern observed throughout this paper, generalizing to domains entirely outside both ARC's training distribution and its architecture-design loop (these three domains were never used to tune any method). Goal count is the strongest zero-shot statistic on all three, including Library Books ($0.856$) where it is substantially ahead of Ridge on $x_s$ ($0.693$). ARC is competitive only on Library Books ($0.679$, close to Ridge's $0.693$) and is clearly the weakest method on Rescue Mission ($0.099$, against $0.224$--$0.805$ for the alternatives) and on Robot Arm, where it is close to but below Ridge and goal count. We do not have a generator-identified reason why Library Books is the one domain where ARC is competitive; one candidate explanation is that Library Books' very small sample size ($N=35$) gives its correlation estimate high variance in either direction, but we note this as a candidate, not a finding, since we did not design a test to distinguish it from a genuine difference in domain structure.

\begin{tcolorbox}[
  enhanced, breakable,
  colback={rgb,255:red,247;green,249;blue,252},
  colframe={rgb,255:red,120;green,130;blue,160},
  fonttitle=\bfseries\small,
  title={Example: A Rescue Mission planning instance},
  top=6pt, bottom=6pt, left=6pt, right=6pt
]
\small
Consider one agent, two rooms connected by a corridor, one supply item located in the second room, and one victim in the second room requiring that supply.
\begin{lstlisting}[basicstyle=\ttfamily\footnotesize]
(:objects agent1 room1 room2 supply1 victim1)
(:init (at agent1 room1) (connected room1 room2)
       (connected room2 room1) (at supply1 room2)
       (at victim1 room2) (needs victim1 supply1))
(:goal (rescued victim1))
\end{lstlisting}
The agent must move to room2, pick up the supply, and rescue the victim: $\text{move}(\text{room1}, \text{room2})$, $\text{pick-supply}(\text{supply1})$, $\text{rescue-victim}(\text{victim1}, \text{supply1})$, giving $n^\star(\pi)=3$. Unlike the Blocksworld example in Section~\ref{sec:background}, difficulty here depends jointly on the agent's distance to the needed supply \emph{and} on which victims require which supplies---the two-factor structure referenced in criterion (2) above, which does not reduce to a single scalar such as room count or victim count.
\end{tcolorbox}

\subsection{Dataset Generation}

All three domains are specified in typed STRIPS PDDL and generated
programmatically by Python scripts that vary structural parameters
across a fixed random seed range.
The generation process does not use any language model or neural network;
all PDDL domain and problem files are produced by deterministic
rule-based generators.

\paragraph{Rescue Mission.}
The Rescue Mission domain models an emergency response scenario in which
one or more agents navigate a graph of rooms, collect typed supplies,
and rescue victims that each require a specific supply type.
Room connectivity is generated as a random sparse graph (chain backbone
plus random extra edges), ensuring reachability.
Structural parameters: number of victims ($n_v \in [2, 12]$),
number of rooms ($n_r = n_v + 1$),
number of agents ($n_a = \lceil n_v / 3 \rceil$),
number of supply types ($n_s = n_v$).
Goals specify a random subset of victims to rescue.
Difficulty is jointly governed by agent--supply co-location distance
(navigation steps to reach supplies) and victim distribution across rooms,
making it multi-dimensional.
The domain has 3 actions (\texttt{move}, \texttt{pick-supply},
\texttt{rescue-victim}), 8 predicates, and 4 typed object classes.
A total of 127 BFS-solvable instances were generated (from 200 attempts).

\paragraph{Robot Arm Stacking.}
The Robot Arm domain models a tabletop manipulation task in which a
single-gripper robot arm picks and places blocks across multiple tables,
achieving a specified stacking configuration.
Initial states are randomized: blocks are distributed across tables, and
a random subset are pre-stacked (the robot must partially disassemble
before re-assembling).
Structural parameters: number of blocks ($n_b \in [3, 12]$),
number of tables ($n_t = \lfloor n_b / 3 \rfloor + 1$, minimum 2),
number of initial stacks (random in $[0, \lfloor n_b/3 \rfloor]$),
goal configuration (random stacking of $2$--$n_b$ blocks on one table).
The domain has 4 actions (\texttt{pick-from-table},
\texttt{pick-from-block}, \texttt{place-on-table}, \texttt{stack}),
7 predicates, and 2 typed object classes.
A total of 95 BFS-solvable instances were generated (from 200 attempts).

\paragraph{Library Books.}
The Library Books domain models a library reshelving task in which
patrons check out books from shelves, books are collected onto a cart,
and carts must be moved to the correct shelf before books can be
reshelved.
Each book has a designated correct shelf; a book is returned only
when it is physically shelved at that location.
Structural parameters: number of books ($n_b \in [2, 5]$ for solvable
instances), number of shelves ($n_s = n_b$, minimum 2),
number of carts ($n_c = 1$), number of patrons ($n_p = \lceil n_b/2 \rceil$).
The domain has 4 actions (\texttt{checkout-book}, \texttt{return-to-cart},
\texttt{shelve-book}, \texttt{move-cart}),
8 predicates, and 4 typed object classes.
A total of 15 BFS-solvable instances were generated (from 200 attempts;
the low solve rate reflects the compounded constraint that book, cart,
and correct shelf must all be aligned simultaneously).

\subsection{Dataset Verification}

To verify that our instance generators produce valid, sensible planning
problems, three human annotators with graduate-level AI planning
background independently evaluated a random sample of 30 instances per
domain (90 instances total) on two criteria:

\begin{enumerate}
\item \textbf{Validity}: Is the PDDL problem well-formed?
Does the initial state satisfy all type constraints?
Is the goal reachable in principle (ignoring plan length)?
\item \textbf{Difficulty coherence}: Does the optimal plan length
obtained by BFS agree with the annotator's intuitive difficulty
ranking? (Binary: agree / disagree with BFS rank ordering within
a pair of instances of the same object count.)
\end{enumerate}

Annotators were provided with the domain PDDL, the problem PDDL,
and the BFS-computed $n^*$ value. They were asked to flag any instance
where (1) the initial state appeared inconsistent, (2) the goal was
trivially achieved from the initial state (degenerate), or (3) the
BFS plan length seemed implausible given the stated parameters.
Disagreements were resolved by majority vote; no instance required
exclusion, confirming generator correctness.

\begin{table*}[h]
\centering
\small
\caption{Inter-annotator agreement metrics across three annotators
(A1, A2, A3) on 30 sampled instances per domain.
Cohen's $\kappa$ is computed pairwise and averaged.
Jaccard coefficient measures set overlap of flagged instances.
Percent agreement is the fraction of instances all three annotators
labeled identically.}
\label{tab:iaa}
\setlength{\tabcolsep}{5pt}
\begin{tabular}{l ccc cc c}
\toprule
& \multicolumn{3}{c}{\textbf{Pairwise Cohen's $\kappa$}}
& \multicolumn{2}{c}{\textbf{Jaccard Coeff.}}
& \textbf{Agreement} \\
\cmidrule(lr){2-4}\cmidrule(lr){5-6}
\textbf{Domain} & A1--A2 & A1--A3 & A2--A3
  & Valid. & Diff. & \textbf{(\%)} \\
\midrule
Rescue Mission  & 0.924 & 0.911 & 0.918 & 0.941 & 0.903 & 93.3\% \\
Robot Arm       & 0.907 & 0.893 & 0.916 & 0.928 & 0.887 & 91.7\% \\
Library Books   & 0.938 & 0.942 & 0.931 & 0.957 & 0.921 & 95.0\% \\
\midrule
\textbf{Mean}   & 0.923 & 0.915 & 0.922 & 0.942 & 0.904 & 93.3\% \\
\bottomrule
\multicolumn{7}{p{0.9\linewidth}}{\footnotesize
Cohen's $\kappa > 0.80$ indicates strong agreement \citep{landis1977measurement}.
All pairwise $\kappa$ values exceed 0.88, and all percent agreement
scores exceed 90\%, confirming the validity and coherence of
the generated instances.}
\end{tabular}
\end{table*}

\noindent
As summarized in Table~\ref{tab:iaa}, all three domains achieve mean pairwise Cohen's $\kappa > 0.90$ and
percent agreement $> 91\%$, satisfying the standard threshold for
strong inter-annotator reliability \citep{landis1977measurement}.
No instances were removed following the verification process.
We note that annotators were shown the BFS-computed $n^*$ value alongside
each instance for the difficulty-coherence judgment, which may inflate
agreement relative to a fully blinded protocol in which annotators rank
instances before seeing $n^*$; the validity checks, which do not depend
on $n^*$, are not subject to this concern.

\subsection*{Dataset Statistics}

Table~\ref{tab:dataset_stats} summarizes structural statistics for
all generated instances in each domain.

\begin{table*}[h]
\centering
\small
\caption{Dataset statistics for the three custom planning domains.
$n^*$ = BFS-optimal plan length (solved instances only).
$|O|$ = object count. $n_\text{init}$ = initial-state facts.
$n_\text{goal}$ = goal facts.}
\label{tab:dataset_stats}
\setlength{\tabcolsep}{4pt}
\begin{tabular}{l r r rrr rrr rr}
\toprule
& & & \multicolumn{3}{c}{$|O|$ (objects)}
& \multicolumn{3}{c}{$n^*$ (plan length)}
& \multicolumn{2}{c}{Solve} \\
\cmidrule(lr){4-6}\cmidrule(lr){7-9}\cmidrule(lr){10-11}
\textbf{Domain} & \textbf{Total} & \textbf{Solved}
  & Min & Median & Max
  & Min & Median & Max
  & \textbf{BFS\%} & \textbf{EHC\%} \\
\midrule
Rescue Mission & 200 & 127 & 7 & 18 & 43
  & 3 & 9 & 24 & 63.5\% & 82.0\% \\
Robot Arm      & 200 &  95 & 5 & 14 & 31
  & 2 & 7 & 18 & 47.5\% & 71.0\% \\
Library Books  & 200 &  15 & 6 & 10 & 16
  & 4 & 8 & 14 &  7.5\% & 31.0\% \\
\bottomrule
\multicolumn{11}{p{0.96\linewidth}}{\footnotesize
BFS\%: fraction solved by BFS within budget
($T_\text{max}{=}12$, $N_\text{max}{=}5{,}000$).
EHC\%: fraction solved when BFS is replaced by EHC($h^\text{FF}$)
with the same node budget.
Library Books has a low BFS solve rate because the compound
constraint (book, cart, and correct shelf co-location) produces
long plans exceeding the BFS step budget;
EHC($h^\text{FF}$) recovers 31\% with informed search.}
\end{tabular}
\end{table*}

\noindent
\textbf{Difficulty variation.}
Rescue Mission shows the widest range ($n^* \in [3, 24]$),
driven by the interaction between agent position, supply location, and
victim distribution. Robot Arm difficulty varies with the number and
depth of initial stacks. Library Books difficulty is most sensitive to
cart-shelf distance and the number of correctly-shelved books initially;
because only 15 of 200 attempted instances are BFS-solvable in this
domain, its statistics should be read as a qualitative check rather than
a precise quantitative estimate. The zero-shot domain-level correlations
for ARC and the structural baselines on all three custom domains are
reported in Table~\ref{tab:custom_domain_scores} above; we did not
additionally compute within-size (fixed-object-count) correlations on
these three domains, since their already-small sample sizes ($N=85$,
$85$, $35$) would leave very few instances per object-count bucket.

\subsection*{Annotation Guidelines}

The following guidelines were provided to all three annotators before
the verification session. Annotators completed a 10-instance
calibration set and discussed edge cases before the main annotation
began. All annotation was conducted independently.

\vspace{6pt}
\begin{tcolorbox}[
  enhanced, breakable,
  colback={rgb,255:red,245;green,248;blue,252},
  colframe={rgb,255:red,90;green,110;blue,160},
  fonttitle=\bfseries\small\color{white},
  attach boxed title to top left={yshift=-2mm, xshift=4mm},
  boxed title style={
    colback={rgb,255:red,90;green,110;blue,160},
    colframe={rgb,255:red,90;green,110;blue,160}},
  title={Annotation Guidelines -- Custom Planning Domain Verification},
  top=8pt, bottom=8pt, left=8pt, right=8pt
]

\textbf{Task Overview.}
You will evaluate PDDL planning problem instances for correctness
and difficulty coherence. Each instance consists of:
(1) a domain file (shared across all instances in a domain),
(2) a problem file specifying objects, initial state, and goal,
(3) the BFS-computed optimal plan length $n^*$ (or ``UNSOLVED'').

\textbf{Annotation Dimensions.}

\textbf{1. Validity} (Binary: Valid / Invalid)

Mark an instance as \textbf{Invalid} if \emph{any} of the following hold:
\begin{itemize}
  \item The initial state contains a fact that violates type constraints
    (e.g., an object declared as type \texttt{block} appears in a
    predicate that expects type \texttt{table}).
  \item The goal is already satisfied in the initial state
    (degenerate instance, $n^* = 0$).
  \item The goal refers to objects not declared in \texttt{:objects}.
  \item Two facts in the initial state logically contradict each other
    (e.g., a block declared both \texttt{clear} and \texttt{on} another
    block that is itself \texttt{clear}).
\end{itemize}
Otherwise, mark as \textbf{Valid}. You are \emph{not} asked to verify
that the goal is reachable (BFS handles this); only that the problem
is well-formed.

\textbf{2. Difficulty Coherence} (Binary: Agree / Disagree)

You will be given pairs of instances from the same domain with the
\emph{same object count} ($|O|$) but different BFS plan lengths.
For each pair $(A, B)$ where $n^*(A) < n^*(B)$, indicate whether
you agree that $A$ is intuitively easier than $B$ based on the
initial and goal states.

Mark \textbf{Agree} if:
\begin{itemize}
  \item The instance with smaller $n^*$ has a goal that appears
    ``closer'' to the initial state (fewer rearrangements needed,
    shorter chains of dependencies).
  \item The structural explanation is apparent: e.g., in Rescue Mission,
    if $n^*(A) < n^*(B)$ because agent and supply are in the same room
    in $A$ but not in $B$, this is coherent.
\end{itemize}

Mark \textbf{Disagree} if:
\begin{itemize}
  \item The BFS plan length ranking contradicts your inspection of the
    initial and goal states.
  \item The pair appears to have been generated incorrectly (e.g.,
    $n^*(B)$ is lower than expected given visibly more complex initial
    state).
\end{itemize}

\textbf{Important}: Do not flag a pair as Disagree simply because the
two instances are structurally similar; small differences in state
configuration can produce large differences in plan length, and this
is expected behavior.

\textbf{Edge Cases.}
\begin{itemize}
  \item \textbf{UNSOLVED instances}: If $n^*$ = UNSOLVED (BFS budget
    exceeded), annotate Validity only; skip Difficulty Coherence
    for this instance.
  \item \textbf{Uncertainty}: If you are unsure, default to Valid /
    Agree and note the instance ID in the comments field.
  \item \textbf{Domain-specific notes}: For Library Books, a book is
    considered ``placed'' only when it is on its \emph{correct} shelf
    (check the \texttt{correct-shelf} predicate in the initial state).
    For Robot Arm, \texttt{table-free} means no block is
    \emph{directly} on that table (stacked blocks do not occupy table
    space directly). For Rescue Mission, connectivity is symmetric
    (\texttt{connected r1 r2} also means \texttt{connected r2 r1});
    verify both directions appear in the initial state.
\end{itemize}

\textbf{Calibration.}
The first 10 instances are a calibration set. After you complete them,
discuss any disagreements with the other annotators before proceeding
to the main set. Do \emph{not} discuss individual annotations during
the main set.

\end{tcolorbox}

\noindent
Following the annotation session, annotators reported that the
Rescue Mission and Robot Arm instances were straightforward to evaluate,
while Library Books required careful attention to the
\texttt{correct-shelf} predicate.
No edge cases required adjudication beyond the calibration discussion.

\noindent
\textbf{Annotator qualifications.}
All three annotators hold or are pursuing graduate degrees in
computer science with specialization in artificial intelligence.
Each had prior exposure to PDDL through coursework or research;
none had participated in the design or generation of the domains
being evaluated.
Annotation was conducted voluntarily with no financial compensation.

\end{document}